\documentclass[10pt,twocolumn,letterpaper]{article}

    \usepackage{iccv}
    \usepackage{times}
    \usepackage{epsfig}
    \usepackage{graphicx}
    \usepackage{amsmath}
    \usepackage{amssymb}
    \usepackage{multirow}
    \usepackage{float}
    \usepackage{bbm}
    \usepackage{enumitem}
    \usepackage{pgf}
    \usepackage{colortbl}
    \usepackage{xifthen}
    \newcommand{\savefootnote}[2]{\footnote{\label{#1}#2}}
    \newcommand{\repeatfootnote}[1]{\textsuperscript{\ref{#1}}}

    \definecolor{light-gray}{gray}{0.95}
    
    \newcommand{\lightercolor}[3]{% Reference Color, Percentage, New Color Name
        \colorlet{#3}{#1!#2!white}
    }
    
    \definecolor{SASHA}{rgb}{0,0.5,0}
    \definecolor{AMIR}{rgb}{1,0,0}
    \definecolor{BRADLY}{rgb}{0,0,1}
    \definecolor{JITENDRA}{rgb}{1,0,0}
    \definecolor{LEO}{rgb}{1,0,0}
    \definecolor{navigation}{rgb}{0.55686275, 0.82352941, 0.99607843}
    \definecolor{exploration}{rgb}{0.96470588, 0.57254902 ,0.56078431}
    \definecolor{geometric_bg}{rgb}{0.98431373,0.70588235, 0.68235294}
    \definecolor{semantic_bg}{rgb}{0.70196078, 0.80392157, 0.89019608}
    \definecolor{texture_bg}{rgb}{0.8, 0.92156863, 0.77254902}
    \definecolor{green_target}{RGB}{140, 248, 90}
    \definecolor{significant}{RGB}{213, 0, 27}
    \definecolor{seaborn}{RGB}{149,177,201}
    \definecolor{seaborn}{RGB}{149,177,201}
    \definecolor{seabornbluetext}{RGB}{0,0,0}
    \definecolor{seabornred}{RGB}{242,197,201}
    \definecolor{seabornredtext}{RGB}{0,0,0}
    \lightercolor{seaborn}{50}{seabornlight}
    \colorlet{2d_feature}{green!75}
    \colorlet{semantic_feature}{blue!75}
    \colorlet{3d_feature}{red!75}
    \colorlet{geometric_fg}{geometric_bg!80!red}
    \colorlet{semantic_fg}{semantic_bg!80!blue}
    
    \newcommand{\PLH}{{\mkern-2mu\times\mkern-2mu}}
    
    \usepackage[pagebackref=true,breaklinks=true,letterpaper=true,colorlinks,bookmarks=false]{hyperref}
    
    \newcommand{\protectedhref}[2]{\ificcvfinal\href{#1}{#2}\else{#2}\fi}
    
     \iccvfinalcopy % *** Uncomment this line for the final submission
    
    \def\iccvPaperID{2447} % *** Enter the ICCV Paper ID here
    
    \newcommand{\defeq}{\raisebox{-0.15\totalheight}{$\triangleq$}}
    
\begin{document}
    
    %%%%%%%%% TITLE
    \title{Mid-Level Visual Representations Improve \\ Generalization and Sample Efficiency for Learning Visuomotor Policies}
    
    \author{\hspace{-2.0mm}Alexander Sax$^{1, 2}$\;\;\hspace{-1.5mm} Bradley Emi$^{3}$\;\;\hspace{-.5mm}Amir Zamir$^{1,3}$\;\;\hspace{-.5mm}Leonidas Guibas$^{2, 3}$ \;\;\hspace{-1.5mm}Silvio Savarese$^{3}$\;\;\hspace{-.5mm}Jitendra Malik$^{1, 2}$ \vspace{10pt}\\ 
        $^1$ University of California, Berkeley\;\;$^2$ Facebook AI Research\;\;$^3$ Stanford University  \vspace{10pt}\\ 
        \textcolor{blue}{\url{http://perceptual.actor/}\vspace{-9pt}}
    }
    
    \maketitle

    \begin{abstract}
    How much does having \textbf{visual priors about the world} (e.g. the fact that the world is 3D) assist in learning to perform \textbf{downstream motor tasks} (e.g. delivering a package)? We study this question by integrating a generic perceptual skill set (e.g. a distance estimator, an edge detector, etc.) within a reinforcement learning framework|see Fig.~\ref{fig:fig1}. This skill set (hereafter \textbf{mid-level perception}) provides the policy with a more processed state of the world compared to raw images.
    
    We find that using a mid-level perception confers significant advantages over training end-to-end from scratch (i.e. not leveraging priors) in navigation-oriented tasks. Agents are able to generalize to situations where the from-scratch approach fails and training becomes significantly more sample efficient. However, we show that realizing these gains requires careful selection of the mid-level perceptual skills. Therefore, we refine our findings into an efficient \textbf{max-coverage feature set} that can be adopted in lieu of raw images. We perform our study in completely separate buildings for training and testing and compare against state-of-the-art feature learning methods and visually blind baseline policies.

    \end{abstract}

    %%%%%%%%% BODY TEXT
    \vspace{-4mm}
    \section{Introduction}
    \vspace{-2mm}
    
    The renaissance of deep reinforcement learning (RL) started with the Atari DQN paper in which Mnih et al.~\cite{mnih-dqn-2015} demonstrated an RL agent that learned to play video games directly from pixels. Levine et al.~\cite{LevineFDA15} adapted this approach to robotics by using RL for control from raw images|a technique commonly referred to as \emph{pixel-to-torque}. The premise of direct-from-pixel learning poses a number of fundamental questions for computer vision: \emph{are perceptual priors about the world actually necessary for learning to perform robotic tasks?} and \emph{what is the value of computer vision objectives, if all one needs from images can be learned from scratch using raw pixels by RL?} 
    %In this paper, we explore the use of representations driven by vision tasks for learning navigation-based motor policies.
    
    %The basic premise of this approach, pertinent to perception, is: \emph{performing an active task can be effectively learned from scratch directly from images}. 
    
    %This interdisciplinary success has led to remarkable recent progress in RL research and has implications for various other fields, in particular, perception.  
    
    \begin{figure}
        \centering
        \includegraphics[width=\columnwidth]{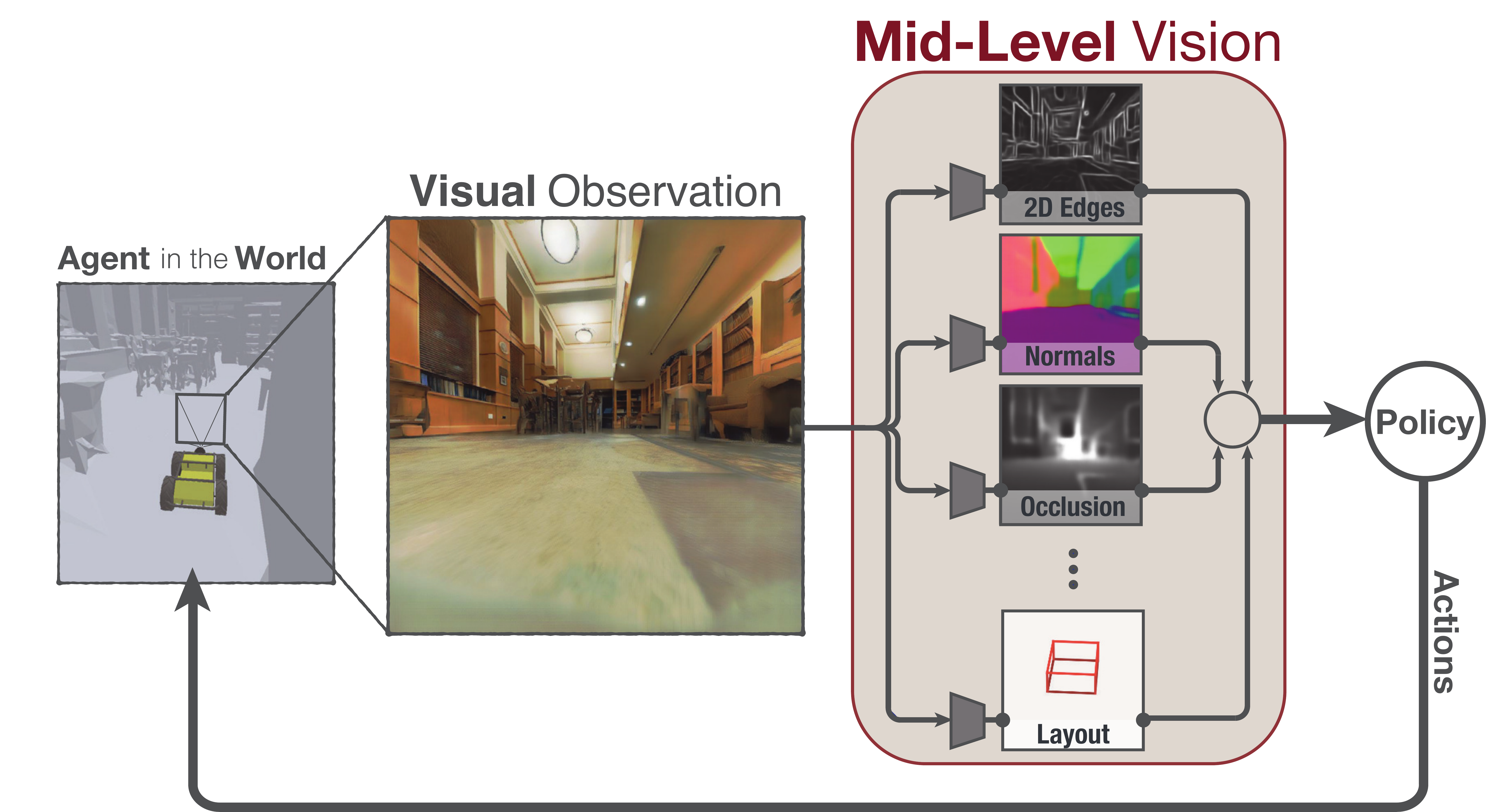}
        \vspace{0mm}
        \caption{\footnotesize{\textbf{A mid-level perception in an end-to-end framework for learning active robotic tasks.} We systematically study if/how a set of generic mid-level vision features can help with learning downstream active tasks. Not incorporating such mid-level perception (i.e. bypassing the red box) is equivalent to learning directly from raw pixels. We report significant advantages in \textit{sample efficiency} and \textit{generalization} when using mid-level perception.}}
        \label{fig:fig1}
    \end{figure}
    
    While deep RL from pixels can learn arbitrary policies in an elegant, end-to-end fashion, there are two phenomena endemic to this paradigm: 
    \textbf{I.} learning requires massive amounts of data (large sample complexity), and
    \textbf{II.} the resulting policies exhibit difficulties reproducing across environments with even modest visual differences (difficulty with generalization). These two phenomena are characteristic of a type of learning that is \emph{overly generic}|in that it does not make use of available \emph{valid assumptions}. 
    Some examples of valid assumptions include that the world is spatially 3D or that certain groupings (``objects'') behave together as a single entity. These are \emph{facts} about the world and are generally true. Incorporating them as priors could provide an advantage over the assumption-free style of learning that always recovers the correct function when given infinite data but struggles when given a limited number of samples~\cite{gemanBiasVariance}. 
    
    %\textbf{II.} the resulting policies exhibit difficulties generalizing across environments with even modest visual differences. These two phenomena are characteristic of a type of learning that is \emph{overly generic}|in that it does not make use of \emph{generally true assumptions}. Some examples of ``generally true'' assumptions are that the world is spatially 3D or certain entities (``objects'') frequently reappear in different places. These are \empth{facts} about the world, and incorporating them as priors in learning yields a useful specialization. Although the assumption-free style of learning always recovers the correct function when given infinite data, it struggles when given a limited number of samples~\cite{gemanBiasVariance}. 

    In this paper, we show that including appropriate perceptual priors can alleviate these two phenomena, improving \emph{generalization} and \emph{sample efficiency}.
    The goal of these priors (more broadly, one of the primary goals of perception) is to provide an internal state that is an understandable representation of the world. In conventional computer vision, this involves defining a set of offline proxy problems (e.g. object detection, depth estimation, etc.) and solving them independently of any ultimate downstream active task~\cite{Codevilla2018offline, Anderson2018evalution}. We study how such standard mid-level vision tasks~\cite{Peirce2015midlevel} and their associated features can be used with RL frameworks in order to train effective visuomotor policies. 
    
    % To be precise, our goal is to train agents to accomplish a \emph{variety} of active tasks in \emph{unseen} test environments. Existing methods often fail when the test setting appears different than the training environment. We study how \emph{mid-level representations}~\cite{Peirce2015midlevel} (e.g. for depth, vanishing points, objects, etc.) could bridge that gap.  
    
    We distill our analysis into three questions: whether these features could improve the: \textbf{I.} \emph{learning speed} (answer: \emph{yes}), \textbf{II.} \emph{generalization to unseen test spaces} (answer: \emph{yes}), and then
    \textbf{III. } \emph{whether a fixed feature could suffice} or a set of features is required for supporting arbitrary motor tasks (answer: \emph{a set is essential}). 
    
    Our findings assert that supporting downstream tasks requires \emph{a set} of visual features. Smaller sets are desirable for both computational efficiency and practical data collection. We put forth a simple and practical solver that takes a large set of features and outputs a smaller feature subset that minimizes the worst-case distance between the selected subset and the best-possible choice. The module can be adopted in lieu of raw pixels to gain the advantages of mid-level vision.
    
    This approach introduces visual biases in a completely computational manner, and it is intermediate between ones that learn everything (like \textit{pixel-to-torque}) and those that leverage fixed models (like classical robotics~\cite{Siciliano2007}). Our study requires learning a visuomotor controller for which we adopted RL|however any of the common alternatives such as control theoretic methods would be viable choices as well.
    In our experiments we use neural networks from existing vision techniques~\cite{taskonomy2018,MITplaces,imagenet,xiazamirhe2018gibsonenv}, trained on real images for specific mid-level tasks. We use their internal representations as the observation provided to the RL policy|we do not use synthetic data to train the visual estimators and do not assume they are perfect. When appropriate, we use statistical tests to answer our questions.
    
    An interactive tool for comparing any trained policies with \href{http://perceptual.actor/policy_explorer/}{videos} and \href{http://perceptual.actor/generalization_curves/}{reward curves}, \href{http://github.com/alexsax/midlevel-reps}{trained models}, and the \href{http://github.com/alexsax/midlevel-reps}{code} are available at \ificcvfinal{at \url{http://perceptual.actor}}\else{\href{/}{URL}}\fi.

    \section{Related Work}
    
    \begin{figure*}
    \vspace{-2mm}
    \includegraphics[width=\textwidth]{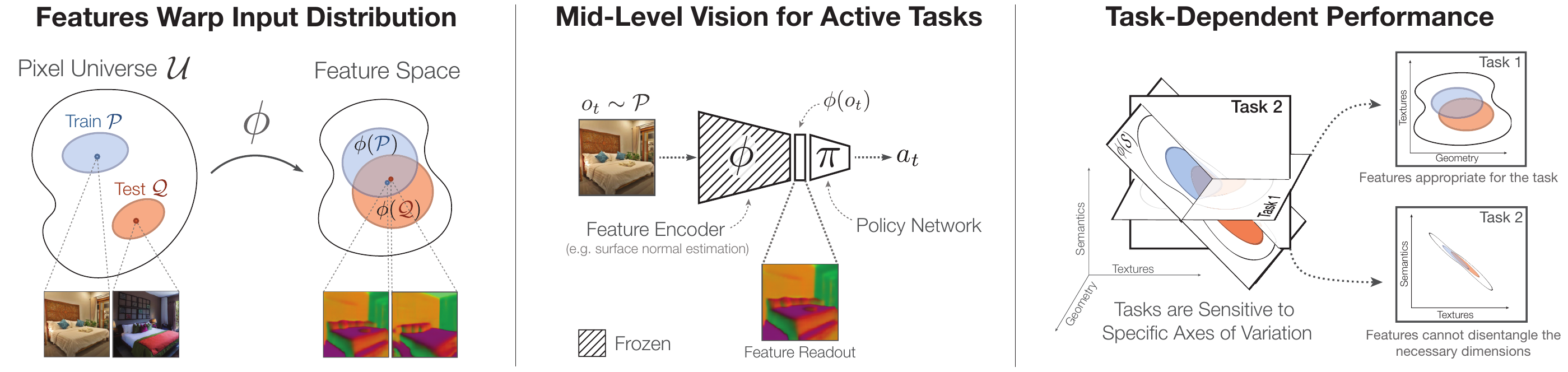}
    % \vspace{-15mm}
       \caption{\footnotesize{\textbf{Illustration of our approach.} Left: Features warp the input distribution, potentially making the train and test distributions look more similar to the agent. Middle: The learned features from fixed encoder networks are used as the state for training policies in RL. Right: Downstream tasks prefer features which contain enough information to solve the task while remaining invariant to the changes in the input which are irrelevant for solving the tasks.}}
           \label{fig:hypothesis_show}
    \end{figure*}

    Our study has connections to a broad set of topics, including lifelong learning, un/self supervised learning, transfer learning, reinforcement and imitation learning, control theory, active vision and several others. We overview the most relevant ones within constraints of space.

    \textbf{Offline Computer Vision} encompasses the approaches
    % |e.g. fully supervised learning~\cite{alexnet}, self-supervised learning~\cite{NorooziF16, pathak2016context, doersch2015unsupervised,zhang2016colorful,amir2016generic,noroozi2017representation}, unsupervised learning~\cite{erhan2010does, bengio2013representation, doersch2015unsupervised,zhang2016colorful,donahue2014decaf,sharif2014cnn}|
    designed to solve various stand-alone vision tasks, e.g. depth estimation~\cite{EigenPF14, laina2016deeper}, object classification~\cite{alexnet, he2016deep}, detection~\cite{yolo9000, Girshick15}, segmentation~\cite{Silberman2012, HeGDG17, Hu2017Segment}, pose estimation~\cite{iss, Cao2018openpose, Xiang2017pose}, etc. The approaches use various levels of supervision~\cite{alexnet, NorooziF16, doersch2015unsupervised, bengio2013representation}, but the common characteristic shared across these methods is that they are \emph{offline} (i.e. trained and tested on prerecorded datasets) and evaluated as a fixed pattern recognition problem. We study how such methods can be plugged into a larger framework for solving downstream active tasks.
    % In recent years the computer vision community has become increasingly interested in robotic tasks. 
    
    \textbf{Reinforcement Learning},~\cite{sutton, MnihKSGAWR13, Silver1140, LevineFDA15} and its variants like Meta-RL~\cite{maml, mamlisbayes18, reptile18, pmaml, bmaml, pmlr-v80-srinivas18b, DuanSCBSA16, MishraRCA17} or its sister fields such as imitation learning \cite{imitation, Giusti2016Imitation, Finn2017imitation, yu2018imitation, KanazawaImitation, RahmatizadehABL17}, commonly focus on the last part of the end-to-end active task pipeline: how to choose an action given a ``state'' from the world. Improvements commonly target the learning algorithm itself (e.g. PPO~\cite{PPO}, Q-Learning~\cite{mnih-dqn-2015}, SAC~\cite{Haarnoja2018softActorCritic}, et cetera), how to efficiently explore the state space~\cite{fu17exploration, curiosity}, or how to balance exploration and exploitation~\cite{agrawal12thompson, Auer2003UCB}. These can be seen as users of our method as we essentially update the input state from pixels (or single fixed features) to a set of generic and presumably more effective vision features. 
    
    \textbf{Representation/Feature Learning} literature shares its goal with our study: how to encode images in a way that provides benefits over using just raw pixels. There has been a remarkable amount of work in this area. They leverage the data either by making some task-agnostic assumption about the data distribution being simpler than the raw pixels (unsupervised approaches like the autoencoder family~\cite{Hinton504, Vincent:2008:ECR:1390156.1390294, kingma2013auto, Matthey2017betaVAELB} and Generative Adversarial Networks~\cite{Szegedy2013adversarial, fu17exploration, Odena2016semisupGan, DonahueKD16}) or they exploit some known structure (so-called \emph{self-supervised} approaches~\cite{zhang2016colorful, noroozi2017representation, NorooziF16, singh2012unsupervised, Wang_2015_ICCV, mousavian18, Gupta17CognitiveMapping, pmlr-v80-srinivas18b, devin2018objcentric, Raffin2019srlbenefits}). A common form of self-supervision in active contexts leverages temporal information to predict unseen observations~\cite{Dosovitskiy16predicting, curiosity, Jordan1992ForwardMS, vandenOord2018predictive}, or actions~\cite{Zhu_2017_ICCV, AgrawalNAML16, curiosity, Raffin2018srltoolbox}. Domain adaptive approaches learn features that are task-relevant but domain agnostic~\cite{Tzeng15adaptation, Saito2017adaptation, TzengHZSD14confusion, Sun16DeepCoral, Long015transferrable, peng18Moment, TzengHSD17Adversarial, bendavid2010adapting}. 
    %Importantly, feature learning involves using a \emph{specific feature} for general tasks or learning a new feature for a \emph{specific task}. When learning \emph{multiple} tasks, the latter approach can not scale, and for the former approach, 
    We show the appropriate choice of feature depends fiercely on the final task. Solving multiple active tasks therefore requires \emph{a set} of features, consistent with recent works in computer vision showing a no single visual feature is the best transfer source for all vision tasks~\cite{taskonomy2018}.
    
    \textbf{Robot Learning} includes methods that leverage fixed models and make hard choices (e.g. which objects are where)~\cite{Siciliano2007} or model-free methods that learn everything from scratch~\cite{LevineEndToEnd15}. These approaches either make assumptions about the world or else require enormous amounts of data, thus they typically work best when restricted to simple domains unrepresentative of the real world (e.g. fixed tabletop domains). For single tasks, methods that leverage some task-specific knowledge in learning have been shown to improve performance~\cite{Chen2019Exploration, Yang2018ScenePriors, Kang2019flying} in realistic environments.
    
    \textbf{Cognitive Psychology} studies~\cite{lake2016building, spelke2007core} suggest that one mechanism for the flexible and sample efficient learning of biological organisms is a universal and evolutionarily ancient set of perceptual biases (such as an object-centric world structure) that tilt learning towards useful visual abstractions. In this paper, we are interesting in embedding a set of perceptual biases into an active artificial agent via a dictionary of mid-level visual features.

    \section{Methodology}
    
    % \[
    % R_\mathcal{Q}(\hat\theta_{P}) = \mathbb{E}_{(s_t, a_t)\sim p_Q(\hat\theta_{P})}[r(s_t, a_t)],
    % \]
    % by estimating the ideal parameters $\hat\theta_\mathcal{P}$ from the training distribution $\mathcal{P}$. 
    
    Our study is focused on agents that maximize the reward in unknown test settings. Our setup assumes access to a set of features $\Phi = \{\phi_1, \ldots, \phi_m\}$, where each feature is a function that can be applied to raw sensory data: transforming the training distribution ($\mathcal{P}$) into $\mathcal{P}_\phi~ \defeq~\phi(\mathcal{P})$ and transforming the test distribution ($\mathcal{Q}$) into $\mathcal{Q}_\phi~ \defeq~\phi(\mathcal{Q})$ (as in Fig.~\ref{fig:hypothesis_show}, left). We examine whether this set can be used to improve this test reward, both in terms of learning speed and generalization (\textit{hypotheses I} and \textit{II} in Sec.~\ref{section:hypotheses}).
    
    \noindent\textbf{Visual Features:}
    \text Figure~\ref{fig:hypothesis_show} shows how a proper feature transforms the training distribution $\mathcal{P}$ into a feature space $\mathcal{P}_\phi$ so that the states at test-time appear similar to those seen during training. In this way, using RL to maximize the training reward also improves the test-time performance, $R_{\mathcal{Q}_\phi}$.
    % \vspace{-2mm}\[R_{\mathcal{Q}_\phi}(\hat\theta_{\mathcal{P}_\phi}) > R_\mathcal{Q}(\hat\theta_{P}). \vspace{-2mm} \]
    Although there are ways~\cite{Tsybakov2008nonparametric} to bound the test performance in terms of the training performance and the shift between the two distributions  $\mathcal{P}_\phi$ and $\mathcal{Q}_\phi$, these bounds are loose in practice and the question of when a feature is helpful remains an empirical one.
    
    % Ideal features need to be invariant to uninformative changes in the environment while maintaining enough discriminability to solve the task (see Fig.~\ref{fig:hypothesis_show}). While there is no \textit{universal} feature (\textit{hypothesis III}, Sec.~\ref{section:hypotheses}), we can choose a minimal covering set of perceptual features (Sec.~\ref{section:selection}) to ensure reasonable generalization on any task.
    
    \subsection{Using Mid-Level Vision for Active Tasks}
    
\begin{figure}
      \vspace{-1mm}
       \includegraphics[width=\columnwidth]{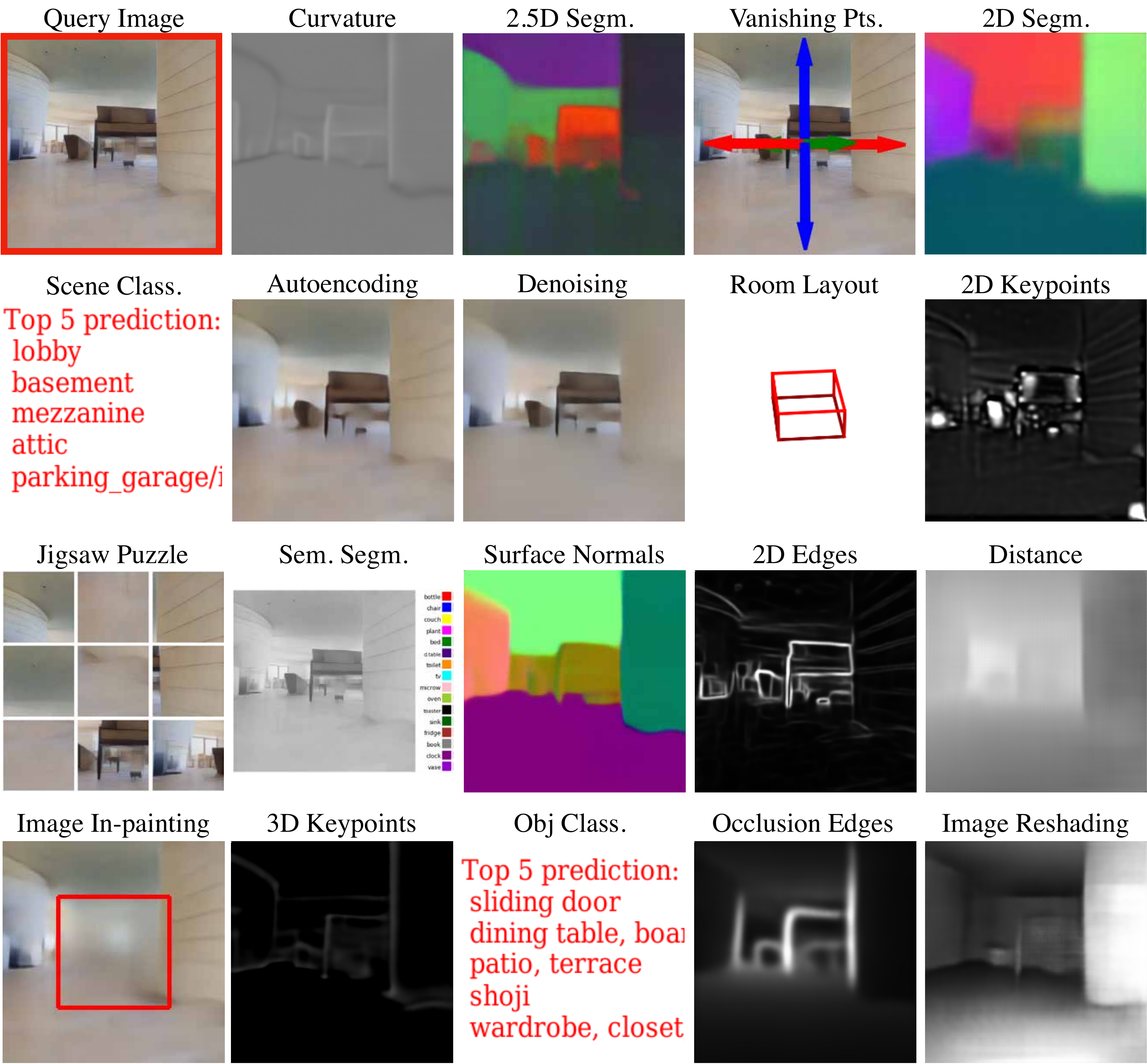}
       \caption{\footnotesize{\textbf{Mid-level vision tasks.} Sample outputs from the vision networks (from Taskonomy~\cite{taskonomy2018} tested on an input from Gibson environment~\cite{xiazamirhe2018gibsonenv}). See more frame-by-frame results on the  \protectedhref{http://perceptual.actor/}{website}.}}
       \label{fig:gibson_feature_representations}
      \vspace{-3mm}
    \end{figure}

    How might we use mid-level perception to support a downstream task? Our mid-level features come from a set of neural networks that were each trained, offline, for a specific mid-level visual task (precisely, 20 networks from~\cite{taskonomy2018} -- see Fig.~\ref{fig:gibson_feature_representations}). We freeze each encoder's weights and use the network ($\phi$) to transform each observed image $o_t$ into a summary statistic $\phi(o_t)$ that we feed to the agent. During training, only the agent policy is updated (as shown in Fig.~\ref{fig:hypothesis_show}, center). Freezing the encoder networks has the advantage that we can reuse the same features for new active tasks without degrading the performance of already-learned policies.
    %, opening the door to learning multiple tasks over the lifetime of the agent. 
    
    % While the features can be optimized for the same environment that the agent will see, our features were trained on standard computer vision datasets. Though the networks are not necessarily well-calibrated in our new environment, we found that the outputs were highly informative (see qualitative examples in Fig. ~\ref{fig:gibson_feature_representations} and the~\protectedhref{http://perceptual.actor/supplementary\#frame_by_frame}{supplementary material}).

    % Of course, other transfer configurations are possible therefore our study provides a lower bound on the benefits of mid-level vision.
    
    % \begin{figure}
    %   \includegraphics[width=\columnwidth]{figures_arxiv19/doom_reps.pdf}
    %   \caption{\footnotesize{\textbf{Feature readouts in Doom.} Sample outputs from the trained perception networks on input from Doom. See more frame-by-frame results in the \protectedhref{http://perceptual.actor/supplementary#frame_by_frame}{supplementary material}.}}
    %   \label{fig:doom_feature_representations2}
    % \end{figure}
    
    \subsection{Core Questions}\label{section:hypotheses}
    
    The following section details our three core hypotheses relating features to agent performance.\\ 
    % \footnote{All RL evaluation suffers from some additional estimation error stemming from the fact that we \emph{estimate} the average performance of a particular random seed and then use that to further estimate the expected performance of a particular training approach. We use enough episodes so that these \textit{cluster effects} are small, but we perform a more sophisticated analysis using~\cite{clusterWilson} in the \protectedhref{http://perceptual.actor/supplementary\#cluster_aware}{supplementary material}, and include the code on our \protectedhref{http://perceptual.actor}{website}}.
    
    \noindent\textbf{Hypothesis I: Sample Efficiency:}
    \textit{Does mid-level vision provide an advantage in terms of sample efficiency when learning an active task?}
    We examine whether an agent equipped with mid-level vision can learn faster than a comparable agent with vision but no visual priors about the world|in other words, an agent learning \textit{tabula rasa}.\\
    
    % Since there is significant variation in the asymptotic performance between different random seeds, any given agent may never achieve a given level of performance. We could not come up with a satisfying metric of relative sample efficiency that could be robustly evaluated from just a few seeds. We therefore provide the curves and ask the readers to use their judgment, which is currently standard practice.
    
    % With our agents defined above, we determine how many interactions with the environment are needed in order to achieve a given level of performance on the test set; we picked 90\% of ``scratch'' best performance as the fixed reference performance, although other ones are viable too and do not alter the conclusions. We check whether any of the features significantly outperform scratch on this metric, correcting for multiple hypothesis testing. 
    
    \noindent\textbf{Hypothesis II: Generalization:}
    \textit{Can agents using mid-level vision generalize better to unseen spaces?}
    If mid-level perception transforms images into standardized encodings that are less environment-specific (Fig.~\ref{fig:hypothesis_show}, left), then we should expect that \emph{agents} using these encodings will learn policies that are more robust to differences between the training and testing environments. We evaluate this in HII by testing which (if any) of the $m$ feature-based agents outperform \textit{tabula rasa} learning in unseen test environments:
    \vspace{-2mm}
    \[ 
    \bigvee_{i=1}^m \big(R_{\mathcal{Q}_{\phi_i}} > R_\mathcal{Q}\big),
    \vspace{-1mm}
    \]
    and correcting for multiple hypothesis testing.\\
    
    \noindent\textbf{Hypothesis III: Single Feature or Feature Set:} \textit{Can a \textbf{single}  feature support all downstream tasks? Or is a set of features required for gaining the feature benefits on arbitrary active tasks?}
    We demonstrate that no feature is universal and can outperform all other features regardless of the downstream activity (represented in the right subplot of Fig.~\ref{fig:hypothesis_show}). We show this by demonstrating cases of \textit{rank-reversal}|when the ideal features for one task are non-ideal for another task (and vice-versa):
    \vspace{-1mm}
    \[
    \big(R^T_{\mathcal{Q}_\phi} > R^T_{\mathcal{Q}_{\phi'}}\big)~\wedge~\big(R^{T'}_{\mathcal{Q}_{\phi'}} > R^{T'}_{\mathcal{Q}_\phi}\big),
    \vspace{-1mm}
    \]
    for tasks $T$ and $T'$ with best features $\phi$ and $\phi'$, respectively.
    
    For instance, we find with high confidence that \emph{depth estimation} features perform well for visual exploration and \emph{object classification} for target-driven navigation, but neither do well vice-versa.

    \subsection{A Covering Set for Mid-Level Perception}\label{section:selection}
    Employing a larger feature set maximizes the change of having the feature proper for the downstream task available. However, a compact set is desirable since agents using a larger set need more data to train|for the same reason that training from raw pixels requires many samples.
    Therefore, we propose a \textbf{Max-Coverage Feature Selector} that curates a compact subset of features to ensure the \emph{ideal} feature (encoder choice) is never too far away from one in the set.

\begin{figure}
    % \centering
    \vspace{-3mm}
    \hspace{2mm}
    \includegraphics[width=1\columnwidth]{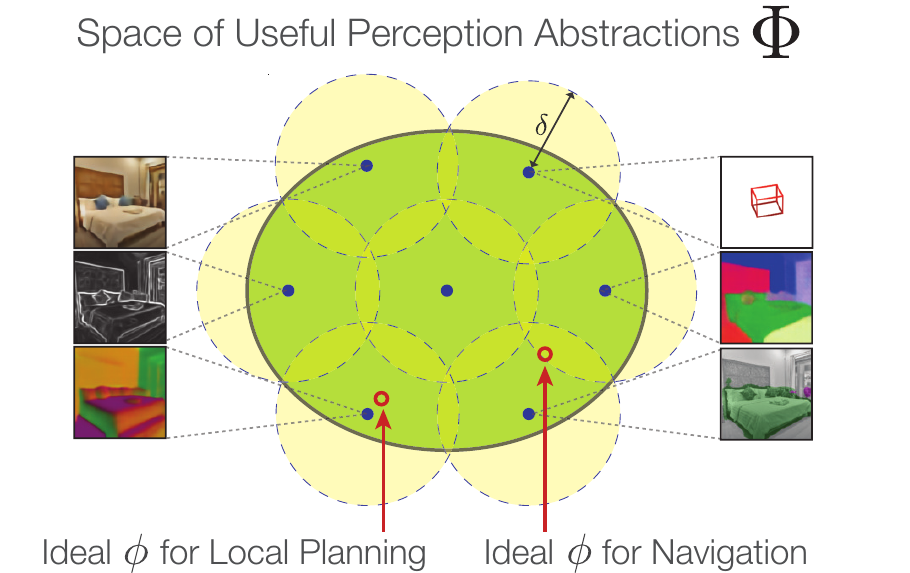}
    \caption{\footnotesize{\textbf{Geometry of the feature set.} We select a covering set of features that minimizes the worst-case distance between the subset and the \emph{ideal} task feature. By \emph{Hypothesis III}, no single feature will suffice and a set is required. okay.}}\label{fig:perception_module}
    \vspace{-1mm}
    % \centering
    % \includegraphics[width=0.8\columnwidth]{figures_arxiv19/perception_covering.eps}
    % % \vspace{-15mm}
    %   \caption{\footnotesize{\textbf{Geometry of the perception module.} Our perception module selects a covering set of features that minimizes the worst-case distance between the set and the \emph{ideal} task feature. By \emph{Hypothesis III}, no single feature will suffice and a set is required.}}
    %       \label{fig:perception_module}
    %     \vspace{-2mm}
    \end{figure}

    The question now becomes how to find the best compact set, shown in Figure~\ref{fig:perception_module}. With a measure of distance between features, we can explicitly minimize the worst-case distance between the best feature and our selected subset (the \emph{perceptual risk} by finding a subset $X_{\delta} \subseteq \Phi = \{\phi_1, ..., \phi_m\}$ of size $|X_{\delta}| \leq k$ that is a $\delta$-cover of $\Phi$ with the smallest possible $\delta$. This is illustrated with a set of size 7 in Figure~\ref{fig:perception_module}.
    
    The task taxonomy method~\cite{taskonomy2018} defines exactly such a distance: a measure between perceptual tasks. Moreover, this measure is predictive of (indeed, derived from) transfer performance. Using this distance, minimizing worst-case transfer (\emph{perceptual risk}) can be formulated as a sequence of Boolean Integer Programs (BIPs)\footnote{For ease of exposition we present a simplified version in the main paper. The full version is similar, but also accounts for feature interactions. See the \protectedhref{http://perceptual.actor/supplementary}{supplementary material}.} parameterized by a boolean vector $x$ indicating which features should be included in the set.
    
        \vspace{-5mm}
        \begin{align*}
        \text{minimize: } & \mathbbm{1}^T x,\\
        \text{subject to: } & Ax \succeq \delta 
        \text{ and } x \in \{0, 1\}^{m}.
        \end{align*}
    Where A is the adjacency matrix of feature distances. That is, the element $a_{ij}$ is the distance from feature $i$ to $j$. This BIP can be solved in under a second.
    
    The above program finds the minimum covering set for any $\delta$. Since there are only $m^2$ distances, we can find the minimum $\delta$ with binary search, by solving $\mathcal{O}(log(m))$ BIPs. This takes under 5 seconds and the final boolean vector $x$ specifies the feature set of size $k$ that minimizes perceptual risk.
    
    \begin{figure*}[ht]
    \vspace{-2mm}
    \includegraphics[width=\textwidth]{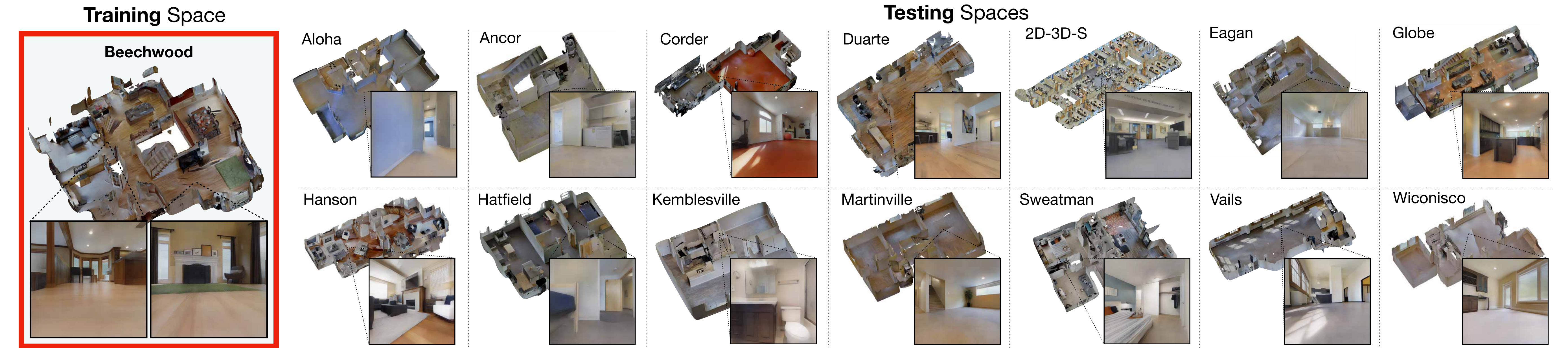}
      \caption{\footnotesize{\textbf{Visualization of training and test buildings from Gibson database~\protect\cite{xiazamirhe2018gibsonenv}.} The training space (on the left, highlighted in red and zoomed) and the testing spaces (remaining on the right). Actual sample observations from agents virtualized in Gibson~\protect\cite{xiazamirhe2018gibsonenv} are shown in the bottom of each box. Results of training and testing in more spaces is provided in the supplementary material.}}
        \label{fig:meshes}
    \end{figure*}
    
    \section{Experiments}
    In this section we describe our experimental setup and present the results from our hypothesis tests and selection module. With 20 vision features and 4 baselines, our approach leads to training between 3-8 seeds per scenario in order to control the false discovery rate~\cite{Benjamini_hochberg}. The total number of policies used in the study is about 800 which took 109,639 GPU-hours to train and evaluate.
    
    \subsection{Experimental Setup}
    \textbf{Environments:} We use the Gibson environment~\cite{xiazamirhe2018gibsonenv} which is designed to be \emph{perceptually} similar to the real world. Training in the real world is difficult due to the intrinsic complexity and reproducibility issues, but Gibson reasonably captures the inherent complexity by virtualizing real buildings and is reproducible. Gibson is also integrated with the PyBullet physics engine which uses a fast collision-handling system to simulate dynamics. We perform our study in Gibson but provide a video of the trained policies tested on real robots in the~\protectedhref{http://perceptual.actor/supplementary\#doom_train_test}{supplementary material}.
    
    % The downside of training in simulation is that there is a realism gap between the simulator and the physical world. We attempt to mitigate this issue in two ways. First, we choose a recent simulator (Gibson~\cite{xiazamirhe2018gibsonenv}) that is designed to be perceptually similar to the real world as it operates by \textit{virtualizing} scans of real buildings. Gibson is also integrated with the PyBullet physics engine which contains a fast collision-handling system used to simulate dynamics.  Second, we perform universality experiments in a second 3D simulator (VizDoom~\cite{vizdoom}) to show that our findings are rather robust to the idiosyncrasies of the particular environment (Sec. \ref{sec:additional_envs}).
    
    \textbf{Train/Test Split:}
    We train and test our agents in two disjoint sets of buildings (Fig.~\ref{fig:meshes}). The test buildings are different and completely unseen during training  The training space for the visual navigation task covers 40.2$m^2$ (square meters) and the testing space covers 415.6$m^2$. For local planning and exploration, the train and test spaces cover 154.9$m^2$ and 1270.1$m^2$.
    % The training space for the local planning and exploration tasks covers 154.9 $m^2$ and the testing space covers 1270.1 $m^2$. 
    % The universality experiments in Doom also use a train/test split of textures which is provided in the \protectedhref{http://perceptual.actor/supplementary\#doom_train_test}{supplementary material}.

    \begin{figure}
    \centering
    \includegraphics[width=1.0\columnwidth]{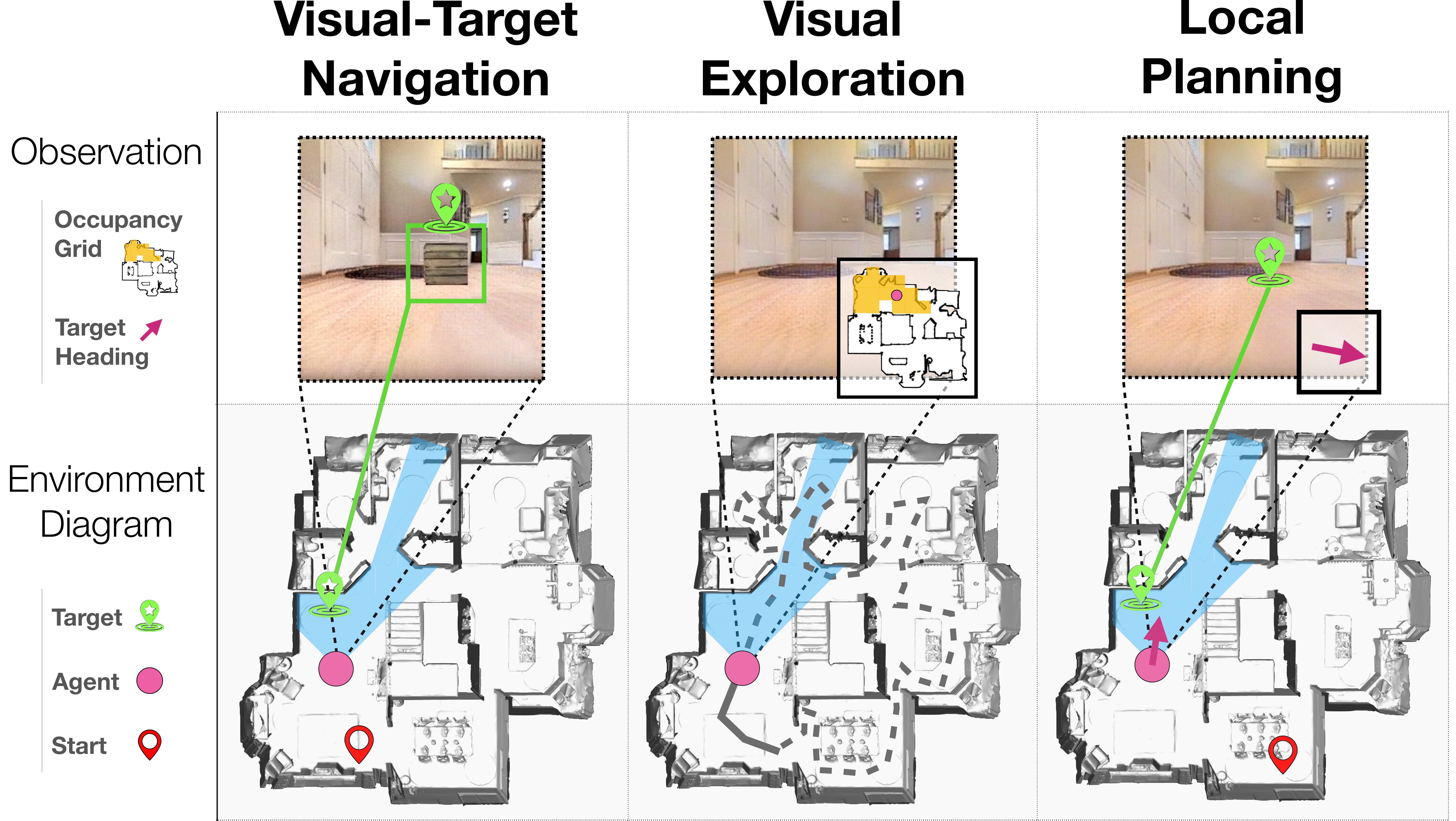}
    \caption{\footnotesize{\textbf{Active task definitions.} Visual descriptions of the selected active tasks and their implementations in Gibson. Additional observations besides the RGB image are shown in the top row. Note that exploration uses only the revealed occupancy grid and no actual mesh boundaries.}} %\sasha{clean up figure, transpose obs, etc.}}}
    \label{fig:task_definition}
    \end{figure}

    % \begin{figure*}[ht]
    % \vspace{-5mm}
    % \hspace{-0mm}
    % % \includegraphics[width=\columnwidth]{figures_arxiv19/trajectory2.pdf}
    % \includegraphics[width=\textwidth]{figures_arxiv19/h1_and_qual_rel_reward_new.pdf}
    %   \caption{\footnotesize{\textbf{Agent trajectories in test environment.} Left: Average rewards in the test environment for features and scratch. Feature-based policies generalize better. Right: Visualizations of the paths that agents take in the test environment. Top: The policy trained for object detection learned to recognize the target and, once it does so, heads for the goal, but fails to cover the entire space in exploration. Middle: Distance estimation features learn some rough approximations for the target, but seems to run around until it is nearly on top of the target, while covering the entire space for exploration. Right: The scratch policy completely fails to generalize to the test space and wanders about almost randomly. More visualizations are available on the \protectedhref{http://perceptual.actor/agent_viz}{website}.}}
    %   \label{figure4}
    % \end{figure*}

    \subsubsection{Downstream Active Tasks}
    In order to test our hypotheses, we sample a few practically useful active tasks: \textit{navigation to a visual target}, \textit{visual exploration}, and \textit{local planning}; depicted in Figure~\ref{fig:task_definition} and described below.
    
    \begin{description}[leftmargin=2mm]
    \small
    \item\textbf{Navigation to a Visual Target:}
    In this scenario the agent must locate a specific target object (a wooden crate) as fast as possible with only \textit{sparse rewards}. Upon touching the target there is a large one-time positive reward (+10) and the episode ends. Otherwise there is a small penalty (-0.025) for living. The target looks the same between episodes although the location and orientation of both the agent and target are randomized according to a uniform distribution over a predefined boundary within the floor plan of the space. The agent must learn to identify the target during the course of training. The maximum episode length is 400 timesteps and the shortest path averages around 30 steps. 
    
    \small
    \item\textbf{Visual Exploration:}
    The agent must visit as many \textbf{new} parts of the space as quickly as possible. The environment is partitioned into small occupancy cells which the agent ``unlocks'' by scanning with a myopic laser range scanner. This scanner reveals the area directly in front of the agent for up to 1.5 meters. The reward at each timestep is proportional to the number of newly revealed cells. The episode ends after 1000 timesteps. 
    %Since our agents are memoryless, we provide them with an odometric map of the unlocked cells.
    
    \small
    \item\textbf{Local Planning:}
    The agent must direct itself to a given nonvisual target destination (specified using coordinates) using visual inputs, avoid obstacles and walls as it navigates to the target. This task is useful for the practical skill of local planning, where an agent must traverse sparse waypoints along a desired path. 
    %Since our agent is memoryless, we keep the problem well-posed by specifying the current target direction.\footnote{This problem formulation is equivalent to assuming the initial coordinates to the target are given, and the robot has a perfect localization system (ideal IMU). In a deployment setting, noise could be added to the target vector to simulate real-world conditions.}
    The agent receives dense positive reward proportional to the progress it makes (in Euclidean distance) toward the goal, and is penalized for colliding with walls and objects. There is also a small negative reward for living as in visual navigation. The maximum episode length is 400 timesteps, and the target distance is sampled from a Gaussian distribution, $~\mathcal{N}(\mu=5~\text{meters}, \sigma^2=2~\text{m)}$.
    %with mean of 5 meters and standard deviation of 2 meters.
    \end{description}

    \textbf{Observation Space:}
    In all tasks, the observation space contains the RGB image and the \emph{minimum} amount of side information needed to feasibly learn the task (Fig.~\ref{fig:task_definition}). Unlike the common practice, we do not include proprioception information such as the agent's joint positions or velocities or any other side information that could be useful, but is not essential to solving the task. We defer the details of each task's observation space to the \protectedhref{http://perceptual.actor/supplementary\#task_descriptions_and_videos}{supplementary material}.
    %We stack the most recent 4 RGB frames as input and do not share weights between these frames to allow the agent to infer its local dynamics. For visual \textbf{navigation}, the state space is only the image. For \textbf{local planning}, the agent also receives the vector to the target in its own inertial reference frame as a vector $[r, \cos{\theta}, \sin{\theta}]$ where $r$ is the Euclidean distance to the target and $\theta$ is the angle relative to the agent's heading in the ground plane\footnote{This problem formulation is equivalent to assuming the initial coordinates to the target are given, and the robot has a perfect localization system (ideal IMU). In a deployment setting, noise could be added to the target vector to simulate real-world conditions.}.  For visual \textbf{exploration}, the task requires some form of memory for the agent to know where it has already been. Since our neural network architecture is memoryless (aside from the frame stacking), we choose to encode the memory as an occupancy grid which is translated and rotated to align with the agent's inertial reference frame, whose cell values are 1 if the agent has already observed a given cell and 0 otherwise. The occupancy grid contains no global information about the scene such as walls or obstacles; it is only the previous output of the robot's laser sensor.

    \textbf{Action Space:}
    We assume a low-level controller for robot actuation, enabling a high-level action space of\vspace{-2mm} $$\mathcal{A}=\small{\{}\texttt{\small{turn\_left}},~\texttt{\small{turn\_right}},~ \texttt{\small{move\_forward}}\small{\}}\vspace{-2mm}.$$
    Detailed specifications can be found in the  \protectedhref{http://perceptual.actor/supplementary}{supplementary material}.

    \subsubsection{Mid-Level Features}
    For our experiments, we used representations derived from one of 20 different computer vision tasks (see Fig.~\ref{fig:gibson_feature_representations}). This set covers various common modes of computer vision tasks: from texture-based (e.g. denoising), to 3D pixel-level (e.g. depth estimation), to low-dimensional geometry (e.g. room layout), to semantic tasks (e.g. object classification). 
    
    We used the networks of~\cite{taskonomy2018} trained on a dataset of 4 million static images in of indoor scenes~\cite{taskonomy2018}. Each network encoder consists of a ResNet-50~\cite{resnet} without a global average-pooling layer. This preserves spatial information in the image. The feature networks were all trained using identical hyperparameters. For a full list of vision tasks, their descriptions, and sample videos of the networks evaluated in our environments, please see the \protectedhref{http://perceptual.actor/}{website}.

    \begin{figure}
    \vspace{-1mm}
    \includegraphics[width=\columnwidth]{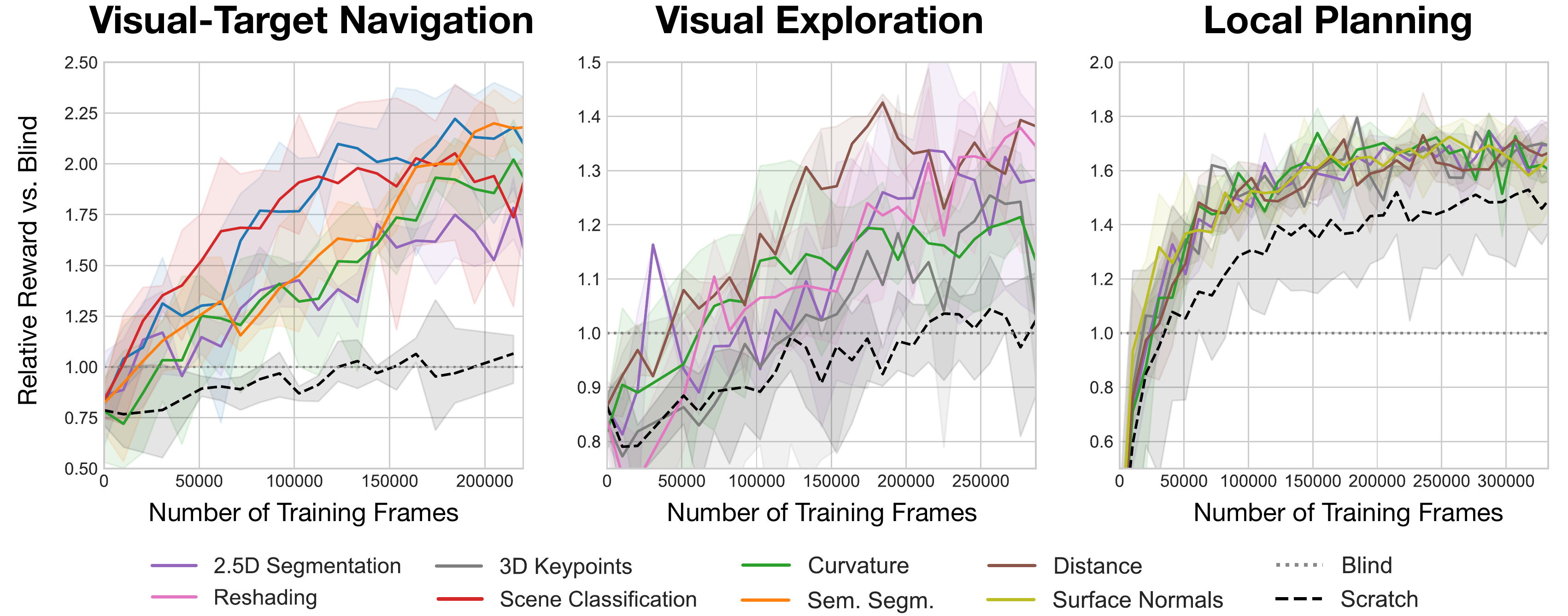}
    \caption{\footnotesize{\textbf{Sample efficiency of feature-based agents.} Average rewards in the test environment for features and scratch. Feature-based policies learn notably faster.
    }}
    \label{fig:h1_curves}
    \end{figure}

    \subsubsection{Reinforcement Learning Algorithm}
    In all experiments we use the common Proximal Policy Optimization (PPO)~\cite{PPO} algorithm with Generalized Advantage Estimation~\cite{gae}. Due to the computational load of rendering perceptually realistic images in Gibson we are only able to use a single rollout worker and we therefore decorrelate our batches using experience replay and off-policy variant of PPO. The formulation is similar to Actor-Critic with Experience Replay (ACER)~\cite{acer} in that full trajectories are sampled from the replay buffer and reweighted using the first-order approximation for importance sampling. We include the full formulation, full experimental details, as well as all network architectures in the \protectedhref{http://perceptual.actor/supplementary}{supplementary material}. 
    
    For each task and each environment we conduct a hyperparameter search optimized for the \textit{scratch} baseline (see section~\ref{section:baselines}). We then fix this setting and reuse it for every feature. This setup favors \textit{scratch} and other baselines that use the same architecture, yet the features outperform them.

    \begin{figure}[]
    \vspace{-4mm}
    \hspace{-0mm}
    \includegraphics[width=\columnwidth]{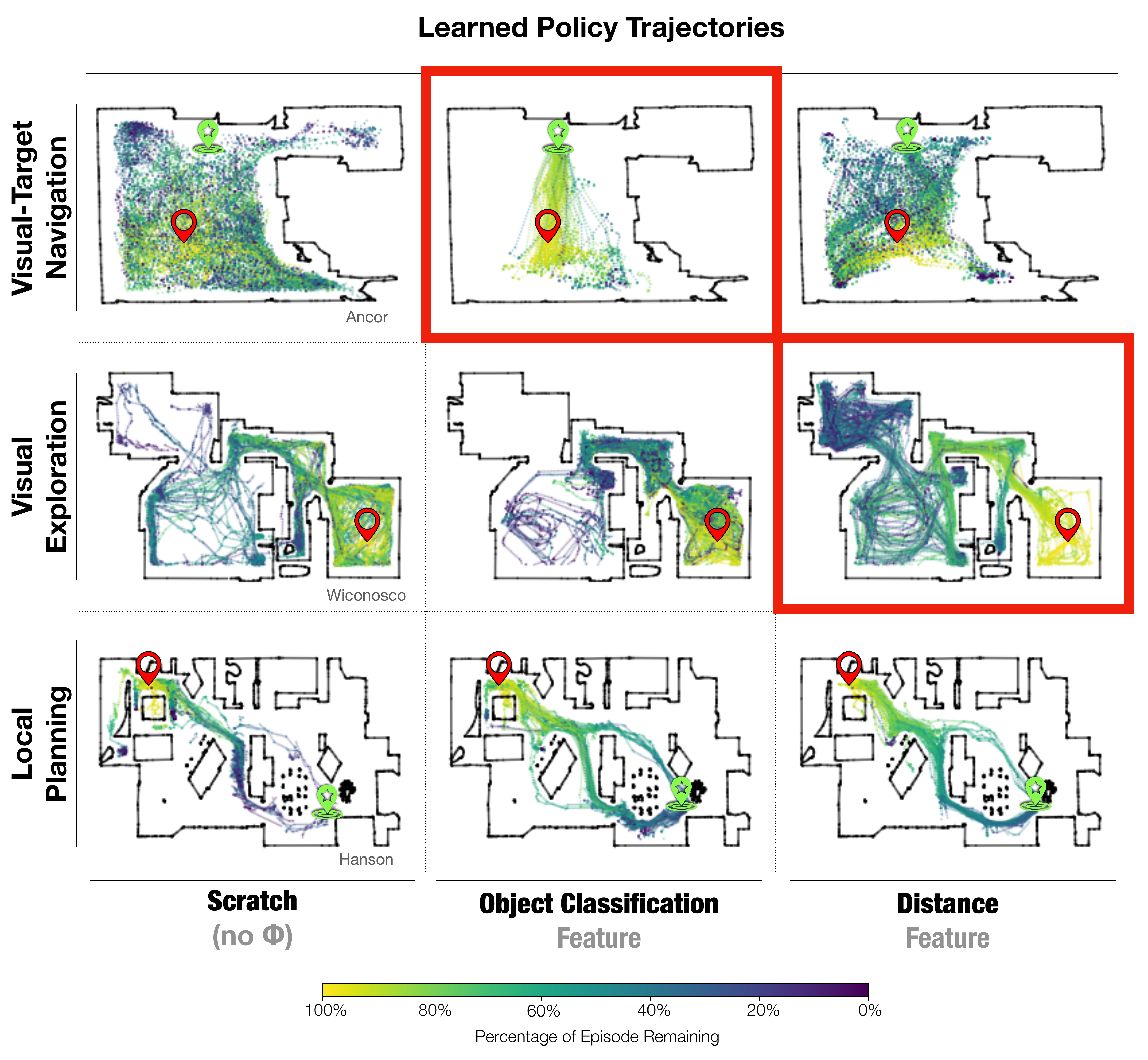}
       \caption{\footnotesize{\textbf{Agent trajectories in test environment.} Left: The \emph{scratch} policy fails to generalize, inefficiently wandering around the test space. Center: The policy trained with \emph{object classification} recognizes and converges on the navigation target (boxed), but fails to cover the entire space in exploration. Right: \emph{Distance estimation} features only help the agent cover nearly the entire space in exploration (boxed), but fail in navigation unless the agent is nearly on top of the target. Visualizations from all features on all tasks are available on the \protectedhref{http://perceptual.actor/policy_explorer/}{website}.}}
       \label{figure4}
    \end{figure}

    \subsection{Baselines}\label{section:baselines}
    
    We include several control groups as baselines which address possible confounding factors:
    
    \begin{description}[leftmargin=2mm]
    \small{
    \item \textbf{Tabula Rasa (Scratch) Learning:}
    The most common approach, \emph{tabula rasa} learning trains the agent from scratch. In this condition (sometimes called \emph{scratch}), the agent receives the raw RGB image as input and uses a randomly initialized AtariNet~\cite{mnih-dqn-2015} tower.
    
    \vspace{-1mm}
    \item \textbf{Blind Intelligent Actor:}
    The \emph{blind} baseline is the same as \emph{tabula rasa} except that the visual input is a fixed image and does not depend on the state of the environment. A \textit{blind} agent indicates how much performance can be squeezed out of the nonvisual biases, correlations, and overall structure of the environment. For instance, in a narrow straight corridor which leads the agent to the target, there should be a small performance gap between \emph{sighted} and \emph{blind}. The \emph{blind} agent is a particularly informative and crucial baseline.
    
    \vspace{-1mm}
    \item \textbf{Random Nonlinear Projections:}
    this is identical to using \emph{mid-level} features, except that the encoder network is randomly initialized and then frozen. The  policy then learns on top of this fixed nonlinear projection.
    %It addresses the possibility that the ResNet architecture, not the offline perception task, is responsible the features' success. 
    % which contains much of the information in the original image. 
    
    \vspace{-1mm}
    \item \textbf{Pixels as Features:}
    this is identical to using \emph{mid-level} features, except that we downsample the input image to the same size as the features ($16\PLH 16$) and use it as the feature.
    This addresses whether the feature readout network could be an improvement over AtariNet which is used for scratch for tractability. 
    
    \vspace{-1mm}
    \item \textbf{Random Actions:}
    uniformly randomly samples from the action space. If random actions perform well then the there is not much to be gained from learning. 
    
    \vspace{-1mm}
    \item \textbf{State-of-the-Art Feature Learning:} offers a comparison of mid-level visual features against several other (not necessarily vision-centric) approaches. We compare against several state-of-the-art feature methods, including dynamic modeling~\cite{Munk2016Forward, ShelhamerMAD16Inverse, Jordan1992ForwardMS}, curiosity~\cite{curiosity}, DARLA~\cite{higgins2017darla}, and ImageNet pretraining~\cite{alexnet}, enumerated in Figure~\ref{fig:sota_srl}.
    
    %This baseline uses the same policy parameterization as the feature-based networks instead of observing a mid-level representation, \emph{pixels-as-state} observes downsampled $16\PLH 16$ images.
    }
    \end{description}

    \begin{figure*}[ht]
        \vspace{-1mm}
        \hspace{-0.00\textwidth}\includegraphics[width=1.00\textwidth]{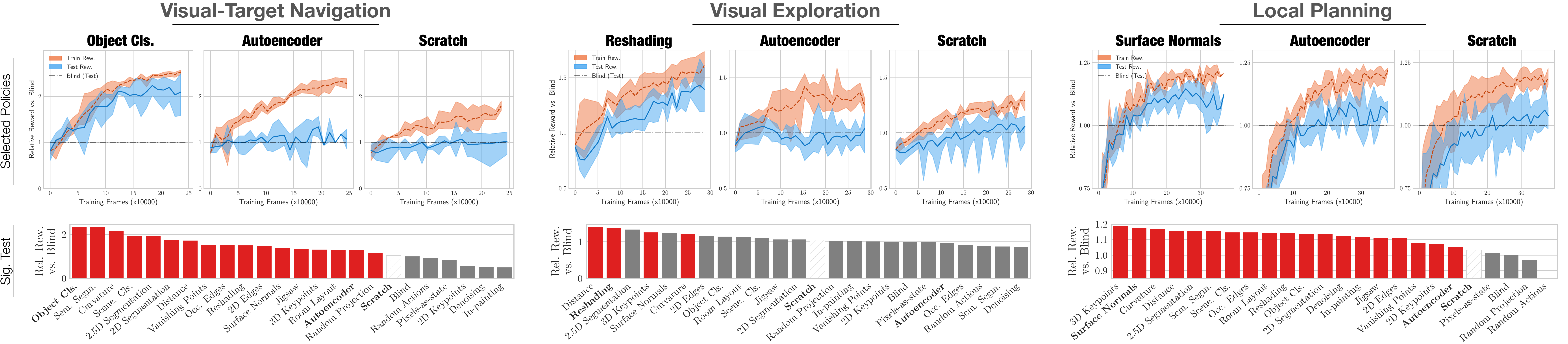}
          \caption{\footnotesize{\textbf{Mid-level feature generalization.} The curve plots above show training and test performance of \textit{scratch} vs. best features throughout training. For all tasks there is a significant gap between train/test performance for \textit{scratch}, and a much smaller one for the best feature. Note that feature-based agents generalize well while \textit{scratch} does not.  \textbf{This underscors the importance of separating the train and test environment in RL}. The bar charts show agent performance in the test environment. Agents \textcolor{significant}{significantly better than \emph{scratch}} are shown in red.}\repeatfootnote{footnote:significance}}
      \label{fig:h2}
     \vspace{-3mm}
    
    \end{figure*}
    
    \subsubsection{Quantification} 
    RL results are typically communicated in terms of absolute reward. However, absolute reward values are uncalibrated and a high value for one task is not necessarily impressive in another. One way to calibrate rewards according to task difficulty is by comparing to a control that cannot access the state of the environment. Therefore, we propose the \textit{reward relative to blind}: \vspace{-2mm}
    \begin{equation}
    RR_{\text{blind}} = \frac{r_\text{treatment} - r_{\text{min}}}{r_\text{blind} - r_{\text{min}}}
    \end{equation}
    as a calibrated quantification. A \emph{blind} agent always achieves a relative reward of 1, while a score $>1$ indicates a relative improvement and score $<1$ indicates this agent performs worse than a \emph{blind} agent. We find this quantification particularly meaningful since we found agents trained from scratch often memorize the training environment, performing no better than \emph{blind} in the test setting (see Fig.~\ref{fig:h2}). We provide the raw reward curves in the \protectedhref{http://perceptual.actor/supplementary}{supplementary material} for completeness.
    
    % [TODO: Test against forwards and backwards dynamics models. Perhaps we should drop this since we are examining perceptual models--not dynamics models. Our methods are compatible with additional forward/inverse dynamics auxilliary losses.]

    \subsection{Experimental results on hypothesis testing I-III}
    \label{sec:hypresults}
    
    In this section we report our findings on the effect of mid-level representations on sample efficiency and generalization. All results are evaluated in the \textit{test} environment with multiple random seeds, unless otherwise explicitly stated. When we use a significance test we opt for a nonparametric approach, sacrificing statistical power to eliminate assumptions on the relevant distributions\savefootnote{footnote:significance}{We use the nonparametric Mann-Whitney U test, correcting for multiple comparisons by controlling the False Discovery Rate ($Q=20\%$) with Benjamini-Hochberg~\cite{Benjamini_hochberg}.}.
    
    \vspace{-1mm}
    \subsubsection{Hypothesis I: Sample Complexity Results}
    
    We find that for each of our active tasks, several feature-based agents learn significantly faster than \emph{scratch}. We evaluated twenty different features against the four control groups on each of our tasks: \textit{visual-target navigation}, \textit{visual exploration}, and \textit{local planning}. Evaluation curves for the five top-performing features appear in Figure~\ref{fig:h1_curves}. Randomly sampled trajectories in Figure~\ref{figure4} highlight how agents trained using features have qualitatively different performance than agents trained \emph{tabula rasa}.

    \vspace{-1mm}
    \subsubsection{Hypothesis II: Generalization Results}\label{sec:h2_analysis}
    
    We find that for each of our tasks, several feature-based agent achieved higher final performance than policies trained \emph{tabula rasa}. We explore conditions when this may \emph{not} hold in Section~\ref{sec:universality}.
    
    \textbf{Large-Scale Analysis:}
    On each task, some features outperform \emph{tabula rasa} learning. Figure~\ref{fig:h2} shows features that outperform \emph{scratch} and those that do so with high confidence are highlighted in red. Significance tests\repeatfootnote{footnote:significance} reveal that the probability of so many results being due to noise is $< 0.002$ per task ($< 10^{-6}$ after the analysis in Sec.~\ref{sec:h3_analysis}).

    \textbf{Mind the Gap:} All agents exhibited some gap between training and test performance, but agents trained from scratch seem to overfit completely---rarely doing better than blind agents in the test environment. The plots in Figure~\ref{fig:h2} show representative examples of this disparity over the course of training. Similarly, some common features like \textit{Autoencoders} and \emph{VAEs} have strong training curves that belie exceptionally weak test-time performance.
    
    %We argue for a more consistent use of train/test splits in order to avoid such illusory victories.
    
    % We show qualitative results in ~\ref{figure4} highlighting the differences between the trajectories taken in both task between an agent trained with a geometric feature (distance estimation) and a semantic feature (object classification), as well as the agent trained from scratch. It is clear that the agent trained with semantic features is able to take a much more efficient route to the goal in the navigation task, since it is immediately able to recognize the target even when displaced in a new environment. However, the semantic agent is much worse at navigating through hallways and doors, as displayed in the exploration task. This is where the depth agent succeeds: despite not having fine grained knowledge about the objects in the scene, as is clear in the navigation task, it can cover much more ground in the exploration task by seeking paths that lead to more wide open spaces. Both tasks perform noticeably better than scratch in the test environment. Since the scratch agent is limited to only the data available during training, it overfits to the specific details of the training environment and cannot generalize.\\
    
    \begin{figure}
        \centering
        \resizebox{1.1\columnwidth}{!}{  
        \hspace{-0.17\columnwidth}\input{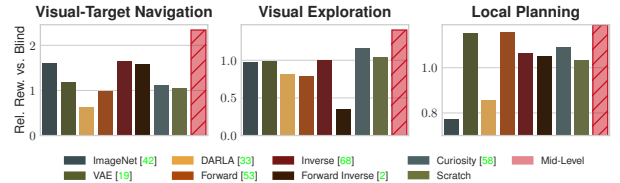}
        }
        \caption{\footnotesize{\textbf{Comparison between our mid-level vision and state-of-the-art feature learning methods}. The vertical axis shows the achieved reward. Note the large gap between mid-level features and the alternatives.}}
        \label{fig:sota_srl}
    \end{figure}
    
    \subsubsection{Hypothesis III: Rank Reversal Results}\label{sec:h3_analysis}
    %Features trained on ImageNet often transfer well and in many cases this is the default choice. However, 
    We found that there may not be one or two single features that consistently outperform all others. Instead, the choice of pretrained features should depend upon the downstream task. This experiment demonstrates this dependence by exhibiting a case of \emph{rank reversal}.
    
    % \begin{figure}
    % \includegraphics[width=\columnwidth]{figures_arxiv19/rank_reversal.pdf}
    % \caption{\footnotesize{\textbf{Rank reversal significance graphs.} Arrows graphs indicate which features are better for a given downstream task. Heavier arrows indicate more significant results (lower $\alpha$-level). \textcolor{navigation}{Blue arrows} point toward tasks that are better for navigation and \textcolor{exploration}{red arrows} point toward tasks that better support exploration. Lack of an arrow indicates the performance difference was not statistically significant. That there is no node with all incoming arrows demonstrates the lack of a universal feature. The essentially complete bipartite structure in the Gibson graph shows that navigation is characteristically semantic while exploration is geometric.}}
    % \label{fig:rank_reversal_graph}
    % \end{figure}
    
    \textbf{Case Study:}
    The top-performing exploration agent used \emph{Distance Estimation} features, perhaps because an effective explorer needs to identify doorways to new, open spaces. In contrast, the top navigation agent used \emph{Object Classification} features---ostensibly because the agent needs to identify the target crate. Despite being top of their class on their preferred tasks, neither feature performed particularly well on the other task.  This result was statistically significant (in both directions) at the $\alpha=0.0005$ level. Fig.~\ref{figure4} visualizes these difference by plotting randomly sampled agent trajectories in a test environment.
    
\begin{figure}
    \vspace{-2mm}
    \includegraphics[width=1.0\columnwidth]{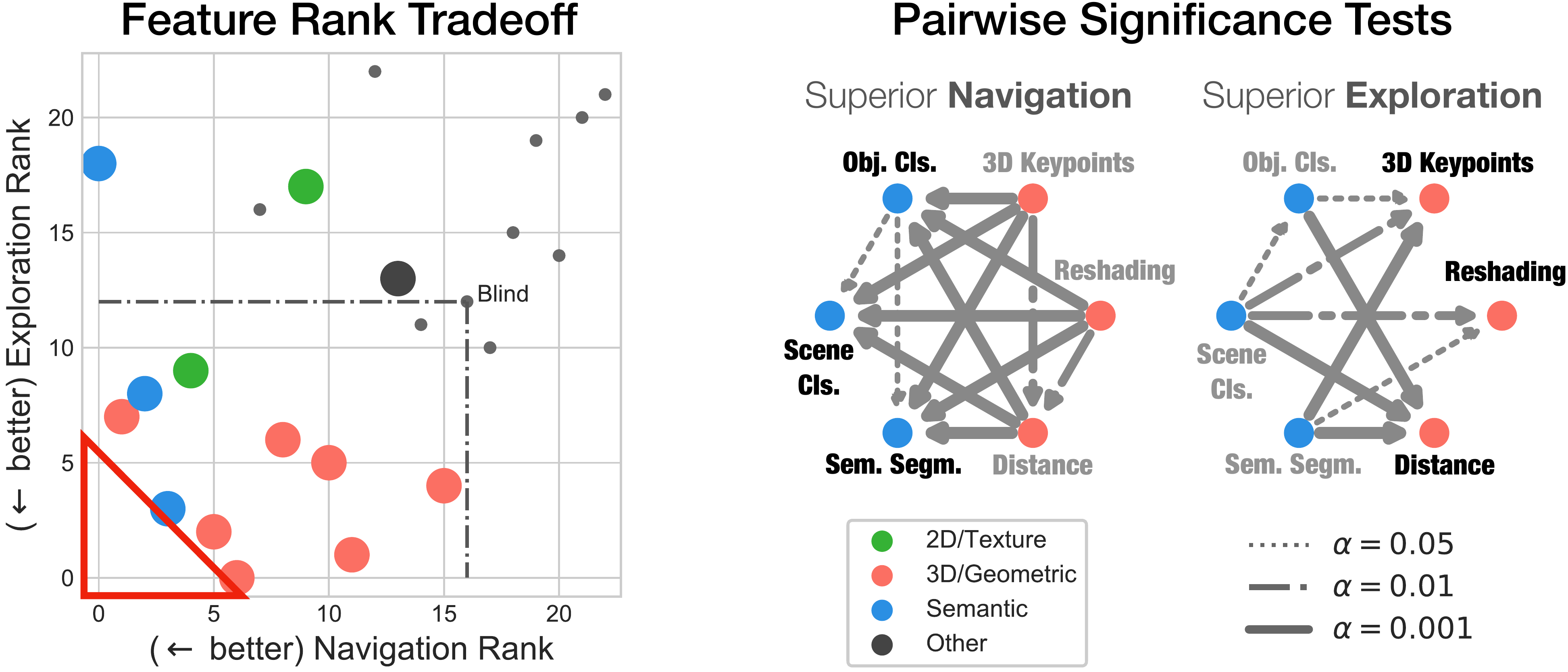}
    \caption{\footnotesize{\textbf{Rank reversal in visual tasks and absence of universal features.} Right: Scatter plot showing feature ranks in navigation (x-axis) and exploration (y-axis). The fact that there is no feature on the bottom left corner (marked with the \textcolor{red}{red triangle}) indicates there was no universal feature. In fact, there is no \emph{almost} universal feature, and maximizing F-score requires giving up 3-4 ranks for each task. Left: 60 pairwise significance tests between the three best features on each task quantify this result. Arrows represent a significant result and they point towards the feature that performed better on the downstream task. \textbf{Heavier arrows} denote higher significance (lower $\alpha$-level). Lack of an arrow indicates that performances were statistically indistinguishable. The essentially complete bipartite structure in the graphs shows that navigation is characteristically semantic while exploration is geometric.}}
    \label{fig:universality_scatter}
    \vspace{-1mm}
    \end{figure}

    \textbf{Ubiquity of Rank Reversal:}
    The trend of rank reversal appears to be a widespread phenomenon. Fig.~\ref{fig:universality_scatter} shows the results from sixty pairwise significance tests, revealing that semantic features are useful for navigation while geometric features are useful for exploration, and that the semantic/geometric distinction is highly predictive of final performance.
    %That rank reversal applies (coarsely) at the family level echoes the findings of~\cite{taskonomy2018}.
    Figure~\ref{fig:sota_srl} shows that state-of-the-art representation learning methods ares similarly task-specific, but the \emph{best} feature outperforms them by a large margin. 
    
    % \amir{the figures sometimes cluster weirdly and irrespective of when they're actually referred to in text. Please check.}
    
    \subsection{Max-Coverage Feature Set Analysis}
    
    % By combining features in a principled way we can improve the performance as much as choosing the best mid-level feature, while introducing less bias than restricting the agent to only a single feature.
    
\begin{figure}
    \vspace{-1mm}
    \includegraphics[width=1.0\columnwidth]{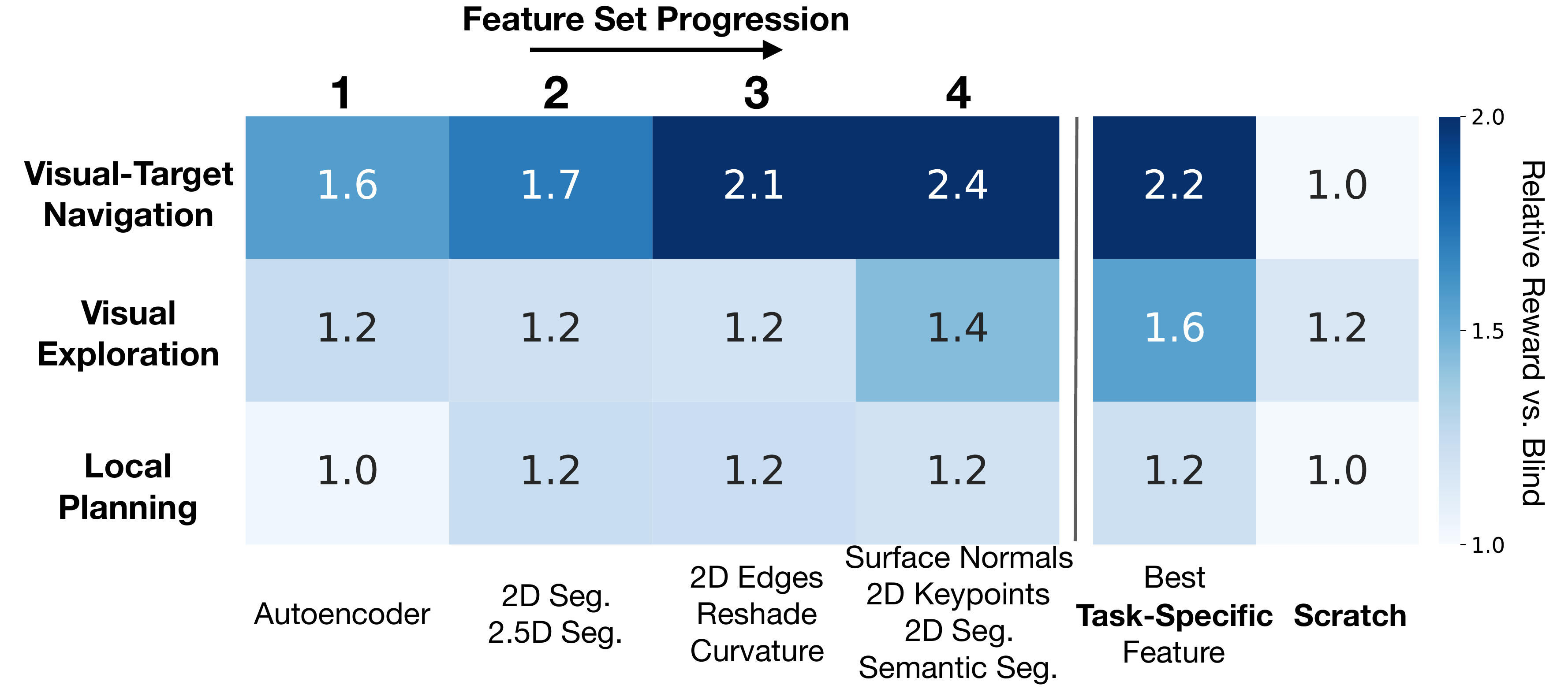}
    \caption{\footnotesize{\textbf{Evaluation of max-coverage feature sets.} The reward (relative to \emph{blind}) is shown in each box. The left four columns show the performance of the agents trained with the max-coverage feature set, as the set size increases from $k$=1 to 4. The two right columns are baselines. The baselines are trained for longer so that all policies in this figure saw the same amount of data.}}
    \label{fig:perception_module_analysis}
    \vspace{-2mm}
    \end{figure}

    The solver described in Section~\ref{section:selection} outputs a set that unifies several useful mid-level vision tasks without sacrificing generality. The experimental results in terms of achieved reward by each feature set (with size $k=1$ to 4) is reported in Figure~\ref{fig:perception_module_analysis}.
    \vspace{-0.5mm}
    \begin{description}[style=unboxed,leftmargin=0cm]
    \vspace{-1.5mm}
    \item\noindent\textbf{Performance:} With only $k=$4 features, our Max-Coverage Feature Set is able to nearly match or exceed the performance of the \emph{best} task-specific feature|even though this set is agnostic to our choice of active tasks.
    
    % \vspace{-1.5mm}
    % \item\noindent\textbf{Feature order:} Earlier, we found that no single feature could perform well on all downstream tasks. However, as number of features in our set increases, so does performance.
    
    \vspace{-1.5mm}
    \item\noindent\textbf{Sample Efficiency:} Due to a larger input space and a larger architecture, we expected worse sample efficiency compared to the best single-feature policy. However, we did not find a noticeable difference.
    
    \vspace{-1.5mm}
    \item\noindent\textbf{Practicality:} This module is able to combine structured sources of information into a compressed representation for use with RL. By simply replacing the raw pixel observation with our max-coverage feature set, practitioners can gain the benefits of mid-level vision. The structure also lends itself well to model-parallelism on modern computational architectures.
    \end{description}
    
    \vspace{-2mm}
    \subsection{Universality Experiments}\label{sec:universality}
    
    \subsubsection{Universality in Additional Buildings}\label{sec:universality_test}
    \vspace{-1mm}
    We repeated our testing in 9 other buildings to account for the possibility that our main test building is anomalous in some way. We found that the reward in our main test building and in the 9 other buildings was extremely strongly correlated with a Spearman's $\rho$ of 0.93 for navigation and 0.85 for exploration. Full experimental setup and results are included in the \protectedhref{http://perceptual.actor/supplementary}{supplementary material}.
    
    % \subsubsection{Universality in Additional Training Buildings}\label{sec:universality_train}
    % In order to rule out the possibility that the training building is in some way unrepresentative of buildings as a whole, we repeated our experiments with agents trained in 7 other buildings (and subsets thereof). The final results were similar and our conclusions were identical. For the complete results, see the \protectedhref{http://perceptual.actor/supplementary}{supplementary material} 
    
    \vspace{-2mm}
    \subsubsection{Universality in Additional Simulators}\label{sec:additional_envs}
    \vspace{-1mm}
    % \begin{figure}
    %     \includegraphics[width=\columnwidth]{figures_arxiv19/doom_generalization_curves_rel_reward.pdf}
    %     \caption{\footnotesize{\textbf{Features generalize to new axes of variation.} In ViZDoom, feature-based agents (right two columns) generalize to new textures even when not exposed to texture variation in training (top row), while agents trained from scratch suffer a significant drop in performance (left, top).}}
    %     \label{fig:doom_texture_generalization}
    % \end{figure}
        
    To evaluate whether our findings are an artifact of the Gibson environment, we tested in an additional environment by implementing navigation and exploration in a second 3D simulator,  VizDoom~\cite{vizdoom}. We found that features which perform well in Gibson also tend to perform well in VizDoom. We also replicated our rank reversal findings (with high confidence), including the geometric/semantic distinction for exploration/navigation and the lack of a \emph{universal feature}. Here, too, maximizing the combined score requires choosing the third- or fourth-best feature for any given task. 
    
    In addition, only feature-based agents were able to generalize without texture randomization during training. Once we added in randomized training textures that resembled the test textures (in effect, making $\mathcal{P}$ and $\mathcal{Q}$ more similar), this distinction disappeared. Our findings do not contradict the general usefulness of RL in the limit of infinite and varied data. Rather, they indicate that to use RL in the (real) world of limited data, we need to introduce learning biases that put their thumb on the scale. The complete VizDoom universality experiments, as well as relevant plots, detailed descriptions of task implementations, and train/test splits are in the  \protectedhref{http://perceptual.actor/supplementary}{supplementary material}.

    \section{Conclusion and Limitations}
    
    This paper presented an approach for using visual biases to learn robotic policies and demonstrated its utility in providing learning biases that improve generalization and reduce sample complexity. We showed that the correct choice of feature depends on the downstream task, and used this fact to refine and generalize our approach: introducing a principled method for selecting a general-purpose feature set. The solver-selected feature sets outperformed state-of-the-art pretraining methods and used at least an order of magnitude less data than learning from scratch|while simultaneously achieving higher final performance. 
    
    A great deal of additional research is possible along this direction. The relationship between visual biases and active tasks is itself an interesting object of study, and a better understanding of these dynamics could lead to more general visual abstractions, as well as ones that are explicitly adapt to specific downstream tasks. In this work, we made a number of simplifying assumptions that are worth noting:
    
    \emph{Locomotive Tasks:} Our selection of active tasks was primarily oriented around locomotion. Though locomotion is a significant enough problem, our study does not necessarily convey conclusions about mid-level vision's utility on other important active tasks, such as manipulation.
    
    \emph{Model Dependence:} We adopted neural networks as our function class. Though
    we validated the stability of our findings on additional environments and against several tasks, in principle our findings could be different if we used another model such as nearest neighbors.
    
    \emph{Reinforcement Learning:} Given that we used RL as our experimental platform, our findings are enveloped by the limitations of existing RL methods, e.g. difficulties in long-range exploration or credit assignment with sparse rewards.
    
    \emph{Lack of Guarantees:} Our approach is primarily empirical. Our measures of agent success and perceptual distance were both derived from experimental results. Successfully predicting agent performance in a test setting would be important for safely deploying robots in a new environment.
    
    \emph{Limited Representation Set:} We used a fixed set of mid-level features, and the best-performing feature necessarily depends on the choice of this set. In addition, we froze the feature weights, limiting how expressive a feature-based policy could possibly be. Relaxing this constraint could improve the worst-case performance to be similar to that of \emph{tabula rasa} learning.
    
    \emph{Lifelong Learning:} Our mid-level feature set is fixed. How to continually update the visual estimators and how to incrementally expand the dictionary are important future research questions.
     
    \vspace{2mm}
    \noindent\textbf{Acknowledgements}
    We gratefully acknowledge the support of ONR MURI~(N00014-14-1-0671), NVIDIA NGC beta, NSF (IIS-1763268), and TRI. Toyota Research Institute (``TRI'')  provided funds to assist the authors with their research but this article solely reflects the opinions and conclusions of its authors and not TRI or any other Toyota entity.
    
    % \todo{other acknowledgements?}

    {\small
    \bibliographystyle{ieee}
    \bibliography{egbib}

\begin{thebibliography}{10}\itemsep=-1pt

\bibitem{imitation}
P.~Abbeel and A.~Y. Ng.
\newblock Apprenticeship learning via inverse reinforcement learning.
\newblock In {\em Proceedings of the Twenty-first International Conference on
  Machine Learning}, ICML '04, pages 1--, New York, NY, USA, 2004. ACM.

\bibitem{AgrawalNAML16}
P.~Agrawal, A.~Nair, P.~Abbeel, J.~Malik, and S.~Levine.
\newblock Learning to poke by poking: Experiential learning of intuitive
  physics.
\newblock {\em CoRR}, abs/1606.07419, 2016.

\bibitem{agrawal12thompson}
S.~Agrawal and N.~Goyal.
\newblock Analysis of thompson sampling for the multi-armed bandit problem.
\newblock In S.~Mannor, N.~Srebro, and R.~C. Williamson, editors, {\em
  Proceedings of the 25th Annual Conference on Learning Theory}, volume~23 of
  {\em Proceedings of Machine Learning Research}, pages 39.1--39.26, Edinburgh,
  Scotland, 25--27 Jun 2012. PMLR.

\bibitem{Anderson2018evalution}
P.~Anderson, A.~X. Chang, D.~S. Chaplot, A.~Dosovitskiy, S.~Gupta, V.~Koltun,
  J.~Kosecka, J.~Malik, R.~Mottaghi, M.~Savva, and A.~R. Zamir.
\newblock On evaluation of embodied navigation agents.
\newblock {\em CoRR}, abs/1807.06757, 2018.

\bibitem{Auer2003UCB}
P.~Auer.
\newblock Using confidence bounds for exploitation-exploration trade-offs.
\newblock {\em J. Mach. Learn. Res.}, 3:397--422, Mar. 2003.

\bibitem{bendavid2010adapting}
S.~Ben-David, J.~Blitzer, K.~Crammer, A.~Kulesza, F.~Pereira, and J.~Vaughan.
\newblock A theory of learning from different domains.
\newblock {\em Machine Learning}, 79:151--175, 2010.

\bibitem{bengio2013representation}
Y.~Bengio, A.~Courville, and P.~Vincent.
\newblock Representation learning: A review and new perspectives.
\newblock {\em IEEE transactions on pattern analysis and machine intelligence},
  35(8):1798--1828, 2013.

\bibitem{Benjamini_hochberg}
Y.~Benjamini and Y.~Hochberg.
\newblock Controlling the false discovery rate: A practical and powerful
  approach to multiple testing.
\newblock {\em Journal of the Royal Statistical Society. Series B
  (Methodological)}, 57(1):289--300, 1995.

\bibitem{Cao2018openpose}
Z.~Cao, G.~Hidalgo, T.~Simon, S.~Wei, and Y.~Sheikh.
\newblock Openpose: Realtime multi-person 2d pose estimation using part
  affinity fields.
\newblock {\em CoRR}, abs/1812.08008, 2018.

\bibitem{Chen2019Exploration}
T.~{Chen}, S.~{Gupta}, and A.~{Gupta}.
\newblock {Learning Exploration Policies for Navigation}.
\newblock {\em arXiv e-prints}, page arXiv:1903.01959, Mar 2019.

\bibitem{Codevilla2018offline}
F.~Codevilla, A.~L{\'{o}}pez, V.~Koltun, and A.~Dosovitskiy.
\newblock On offline evaluation of vision-based driving models.
\newblock {\em CoRR}, abs/1809.04843, 2018.

\bibitem{imagenet}
J.~Deng, W.~Dong, R.~Socher, L.-J. Li, K.~Li, and L.~Fei-Fei.
\newblock {ImageNet: A Large-Scale Hierarchical Image Database}.
\newblock In {\em CVPR09}, 2009.

\bibitem{devin2018objcentric}
C.~{Devin}, P.~{Abbeel}, T.~{Darrell}, and S.~{Levine}.
\newblock Deep object-centric representations for generalizable robot learning.
\newblock In {\em 2018 IEEE International Conference on Robotics and Automation
  (ICRA)}, pages 7111--7118, May 2018.

\bibitem{doersch2015unsupervised}
C.~Doersch, A.~Gupta, and A.~A. Efros.
\newblock Unsupervised visual representation learning by context prediction.
\newblock In {\em Proceedings of the IEEE International Conference on Computer
  Vision}, pages 1422--1430, 2015.

\bibitem{DonahueKD16}
J.~Donahue, P.~Kr{\"{a}}henb{\"{u}}hl, and T.~Darrell.
\newblock Adversarial feature learning.
\newblock {\em CoRR}, abs/1605.09782, 2016.

\bibitem{Dosovitskiy16predicting}
A.~Dosovitskiy and V.~Koltun.
\newblock Learning to act by predicting the future.
\newblock {\em CoRR}, abs/1611.01779, 2016.

\bibitem{DuanSCBSA16}
Y.~Duan, J.~Schulman, X.~Chen, P.~L. Bartlett, I.~Sutskever, and P.~Abbeel.
\newblock Rl{\textdollar}{\^{}}2{\textdollar}: Fast reinforcement learning via
  slow reinforcement learning.
\newblock {\em CoRR}, abs/1611.02779, 2016.

\bibitem{EigenPF14}
D.~Eigen, C.~Puhrsch, and R.~Fergus.
\newblock Depth map prediction from a single image using a multi-scale deep
  network.
\newblock {\em CoRR}, abs/1406.2283, 2014.

\bibitem{Eslami1204}
S.~M.~A. Eslami, D.~Jimenez~Rezende, F.~Besse, F.~Viola, A.~S. Morcos,
  M.~Garnelo, A.~Ruderman, A.~A. Rusu, I.~Danihelka, K.~Gregor, D.~P. Reichert,
  L.~Buesing, T.~Weber, O.~Vinyals, D.~Rosenbaum, N.~Rabinowitz, H.~King,
  C.~Hillier, M.~Botvinick, D.~Wierstra, K.~Kavukcuoglu, and D.~Hassabis.
\newblock Neural scene representation and rendering.
\newblock {\em Science}, 360(6394):1204--1210, 2018.

\bibitem{maml}
C.~Finn, P.~Abbeel, and S.~Levine.
\newblock Model-agnostic meta-learning for fast adaptation of deep networks.
\newblock {\em CoRR}, abs/1703.03400, 2017.

\bibitem{pmaml}
C.~{Finn}, K.~{Xu}, and S.~{Levine}.
\newblock {Probabilistic Model-Agnostic Meta-Learning}.
\newblock {\em ArXiv e-prints}, June 2018.

\bibitem{Finn2017imitation}
C.~Finn, T.~Yu, T.~Zhang, P.~Abbeel, and S.~Levine.
\newblock One-shot visual imitation learning via meta-learning.
\newblock {\em CoRR}, abs/1709.04905, 2017.

\bibitem{fu17exploration}
J.~Fu, J.~D. Co{-}Reyes, and S.~Levine.
\newblock {EX2:} exploration with exemplar models for deep reinforcement
  learning.
\newblock {\em CoRR}, abs/1703.01260, 2017.

\bibitem{gemanBiasVariance}
S.~{Geman}, E.~{Bienenstock}, and R.~{Doursat}.
\newblock Neural networks and the bias/variance dilemma.
\newblock {\em Neural Computation}, 4(1):1--58, Jan 1992.

\bibitem{Girshick15}
R.~B. Girshick.
\newblock Fast {R-CNN}.
\newblock {\em CoRR}, abs/1504.08083, 2015.

\bibitem{Giusti2016Imitation}
A.~{Giusti}, J.~{Guzzi}, D.~C. {Cireşan}, F.~{He}, J.~P. {Rodríguez},
  F.~{Fontana}, M.~{Faessler}, C.~{Forster}, J.~{Schmidhuber}, G.~D. {Caro},
  D.~{Scaramuzza}, and L.~M. {Gambardella}.
\newblock A machine learning approach to visual perception of forest trails for
  mobile robots.
\newblock {\em IEEE Robotics and Automation Letters}, 1(2):661--667, July 2016.

\bibitem{mamlisbayes18}
E.~Grant, C.~Finn, S.~Levine, T.~Darrell, and T.~L. Griffiths.
\newblock Recasting gradient-based meta-learning as hierarchical bayes.
\newblock {\em CoRR}, abs/1801.08930, 2018.

\bibitem{Gupta17CognitiveMapping}
S.~Gupta, J.~Davidson, S.~Levine, R.~Sukthankar, and J.~Malik.
\newblock Cognitive mapping and planning for visual navigation.
\newblock {\em CoRR}, abs/1702.03920, 2017.

\bibitem{Haarnoja2018softActorCritic}
T.~Haarnoja, A.~Zhou, P.~Abbeel, and S.~Levine.
\newblock Soft actor-critic: Off-policy maximum entropy deep reinforcement
  learning with a stochastic actor.
\newblock {\em CoRR}, abs/1801.01290, 2018.

\bibitem{HeGDG17}
K.~He, G.~Gkioxari, P.~Doll{\'{a}}r, and R.~B. Girshick.
\newblock Mask {R-CNN}.
\newblock {\em CoRR}, abs/1703.06870, 2017.

\bibitem{resnet}
K.~He, X.~Zhang, S.~Ren, and J.~Sun.
\newblock Deep residual learning for image recognition.
\newblock {\em CoRR}, abs/1512.03385, 2015.

\bibitem{he2016deep}
K.~He, X.~Zhang, S.~Ren, and J.~Sun.
\newblock Deep residual learning for image recognition.
\newblock In {\em Proceedings of the IEEE conference on computer vision and
  pattern recognition}, pages 770--778, 2016.

\bibitem{higgins2017darla}
I.~{Higgins}, A.~{Pal}, A.~A. {Rusu}, L.~{Matthey}, C.~P. {Burgess},
  A.~{Pritzel}, M.~{Botvinick}, C.~{Blundell}, and A.~{Lerchner}.
\newblock {DARLA: Improving Zero-Shot Transfer in Reinforcement Learning}.
\newblock {\em arXiv e-prints}, page arXiv:1707.08475, Jul 2017.

\bibitem{Hinton504}
G.~E. Hinton and R.~R. Salakhutdinov.
\newblock Reducing the dimensionality of data with neural networks.
\newblock {\em Science}, 313(5786):504--507, 2006.

\bibitem{Hu2017Segment}
R.~Hu, P.~Doll{\'{a}}r, K.~He, T.~Darrell, and R.~B. Girshick.
\newblock Learning to segment every thing.
\newblock {\em CoRR}, abs/1711.10370, 2017.

\bibitem{Jordan1992ForwardMS}
M.~I. Jordan and D.~E. Rumelhart.
\newblock Forward models: Supervised learning with a distal teacher.
\newblock {\em Cognitive Science}, 16:307--354, 1992.

\bibitem{KanazawaImitation}
A.~Kanazawa, J.~Zhang, P.~Felsen, and J.~Malik.
\newblock Learning 3d human dynamics from video.
\newblock {\em CoRR}, abs/1812.01601, 2018.

\bibitem{Kang2019flying}
K.~{Kang}, S.~{Belkhale}, G.~{Kahn}, P.~{Abbeel}, and S.~{Levine}.
\newblock {Generalization through Simulation: Integrating Simulated and Real
  Data into Deep Reinforcement Learning for Vision-Based Autonomous Flight}.
\newblock {\em arXiv e-prints}, page arXiv:1902.03701, Feb 2019.

\bibitem{vizdoom}
M.~Kempka, M.~Wydmuch, G.~Runc, J.~Toczek, and W.~Jaskowski.
\newblock Vizdoom: {A} doom-based {AI} research platform for visual
  reinforcement learning.
\newblock {\em CoRR}, abs/1605.02097, 2016.

\bibitem{bmaml}
T.~{Kim}, J.~{Yoon}, O.~{Dia}, S.~{Kim}, Y.~{Bengio}, and S.~{Ahn}.
\newblock {Bayesian Model-Agnostic Meta-Learning}.
\newblock {\em ArXiv e-prints}, June 2018.

\bibitem{kingma2013auto}
D.~P. Kingma and M.~Welling.
\newblock Auto-encoding variational bayes.
\newblock {\em arXiv preprint arXiv:1312.6114}, 2013.

\bibitem{alexnet}
A.~Krizhevsky, I.~Sutskever, and G.~E. Hinton.
\newblock Imagenet classification with deep convolutional neural networks.
\newblock In F.~Pereira, C.~J.~C. Burges, L.~Bottou, and K.~Q. Weinberger,
  editors, {\em Advances in Neural Information Processing Systems 25}, pages
  1097--1105. Curran Associates, Inc., 2012.

\bibitem{laina2016deeper}
I.~Laina, C.~Rupprecht, V.~Belagiannis, F.~Tombari, and N.~Navab.
\newblock Deeper depth prediction with fully convolutional residual networks.
\newblock In {\em 3D Vision (3DV), 2016 Fourth International Conference on},
  pages 239--248. IEEE, 2016.

\bibitem{lake2016building}
B.~M. Lake, T.~D. Ullman, J.~B. Tenenbaum, and S.~J. Gershman.
\newblock Building machines that learn and think like people.
\newblock {\em Behavioral and Brain Sciences}, pages 1--101, 2016.

\bibitem{LevineFDA15}
S.~Levine, C.~Finn, T.~Darrell, and P.~Abbeel.
\newblock End-to-end training of deep visuomotor policies.
\newblock {\em CoRR}, abs/1504.00702, 2015.

\bibitem{LevineEndToEnd15}
S.~Levine, C.~Finn, T.~Darrell, and P.~Abbeel.
\newblock End-to-end training of deep visuomotor policies.
\newblock {\em CoRR}, abs/1504.00702, 2015.

\bibitem{Long015transferrable}
M.~Long and J.~Wang.
\newblock Learning transferable features with deep adaptation networks.
\newblock {\em CoRR}, abs/1502.02791, 2015.

\bibitem{Matthey2017betaVAELB}
L.~Matthey, A.~Pal, C.~Burgess, X.~Glorot, M.~Botvinick, S.~Mohamed, and
  A.~Lerchner.
\newblock beta-vae: Learning basic visual concepts with a constrained
  variational framework.
\newblock In {\em ICLR 2017}, 2017.

\bibitem{MishraRCA17}
N.~Mishra, M.~Rohaninejad, X.~Chen, and P.~Abbeel.
\newblock Meta-learning with temporal convolutions.
\newblock {\em CoRR}, abs/1707.03141, 2017.

\bibitem{MnihKSGAWR13}
V.~Mnih, K.~Kavukcuoglu, D.~Silver, A.~Graves, I.~Antonoglou, D.~Wierstra, and
  M.~A. Riedmiller.
\newblock Playing atari with deep reinforcement learning.
\newblock {\em CoRR}, abs/1312.5602, 2013.

\bibitem{mnih-dqn-2015}
V.~Mnih, K.~Kavukcuoglu, D.~Silver, A.~A. Rusu, J.~Veness, M.~G. Bellemare,
  A.~Graves, M.~Riedmiller, A.~K. Fidjeland, G.~Ostrovski, S.~Petersen,
  C.~Beattie, A.~Sadik, I.~Antonoglou, H.~King, D.~Kumaran, D.~Wierstra,
  S.~Legg, and D.~Hassabis.
\newblock Human-level control through deep reinforcement learning.
\newblock {\em Nature}, 518(7540):529--533, 02 2015.

\bibitem{mousavian18}
A.~Mousavian, A.~Toshev, M.~Fiser, J.~Kosecka, and J.~Davidson.
\newblock Visual representations for semantic target driven navigation.
\newblock {\em CoRR}, abs/1805.06066, 2018.

\bibitem{Munk2016Forward}
J.~{Munk}, J.~{Kober}, and R.~{Babuška}.
\newblock Learning state representation for deep actor-critic control.
\newblock In {\em 2016 IEEE 55th Conference on Decision and Control (CDC)},
  pages 4667--4673, Dec 2016.

\bibitem{reptile18}
A.~Nichol, J.~Achiam, and J.~Schulman.
\newblock On first-order meta-learning algorithms.
\newblock {\em CoRR}, abs/1803.02999, 2018.

\bibitem{NorooziF16}
M.~Noroozi and P.~Favaro.
\newblock Unsupervised learning of visual representations by solving jigsaw
  puzzles.
\newblock In {\em European Conference on Computer Vision}, pages 69--84.
  Springer, 2016.

\bibitem{noroozi2017representation}
M.~Noroozi, H.~Pirsiavash, and P.~Favaro.
\newblock Representation learning by learning to count.
\newblock {\em arXiv preprint arXiv:1708.06734}, 2017.

\bibitem{Odena2016semisupGan}
A.~{Odena}.
\newblock {Semi-Supervised Learning with Generative Adversarial Networks}.
\newblock {\em arXiv e-prints}, page arXiv:1606.01583, Jun 2016.

\bibitem{curiosity}
D.~Pathak, P.~Agrawal, A.~A. Efros, and T.~Darrell.
\newblock Curiosity-driven exploration by self-supervised prediction.
\newblock {\em CoRR}, abs/1705.05363, 2017.

\bibitem{Peirce2015midlevel}
J.~W. Peirce.
\newblock {Understanding mid-level representations in visual processing}.
\newblock {\em Journal of Vision}, 15(7):5--5, 06 2015.

\bibitem{peng18Moment}
X.~Peng, Q.~Bai, X.~Xia, Z.~Huang, K.~Saenko, and B.~Wang.
\newblock Moment matching for multi-source domain adaptation.
\newblock {\em CoRR}, abs/1812.01754, 2018.

\bibitem{Raffin2019srlbenefits}
A.~Raffin, A.~Hill, K.~R. Traor{\'{e}}, T.~Lesort, N.~D. Rodr{\'{\i}}guez, and
  D.~Filliat.
\newblock Decoupling feature extraction from policy learning: assessing
  benefits of state representation learning in goal based robotics.
\newblock {\em CoRR}, abs/1901.08651, 2019.

\bibitem{Raffin2018srltoolbox}
A.~Raffin, A.~Hill, R.~Traor{\'{e}}, T.~Lesort, N.~D. Rodr{\'{\i}}guez, and
  D.~Filliat.
\newblock {S-RL} toolbox: Environments, datasets and evaluation metrics for
  state representation learning.
\newblock {\em CoRR}, abs/1809.09369, 2018.

\bibitem{RahmatizadehABL17}
R.~Rahmatizadeh, P.~Abolghasemi, L.~B{\"{o}}l{\"{o}}ni, and S.~Levine.
\newblock Vision-based multi-task manipulation for inexpensive robots using
  end-to-end learning from demonstration.
\newblock {\em CoRR}, abs/1707.02920, 2017.

\bibitem{yolo9000}
J.~Redmon and A.~Farhadi.
\newblock {YOLO9000:} better, faster, stronger.
\newblock {\em CoRR}, abs/1612.08242, 2016.

\bibitem{Saito2017adaptation}
K.~Saito, K.~Watanabe, Y.~Ushiku, and T.~Harada.
\newblock Maximum classifier discrepancy for unsupervised domain adaptation.
\newblock {\em CoRR}, abs/1712.02560, 2017.

\bibitem{gae}
J.~Schulman, P.~Moritz, S.~Levine, M.~I. Jordan, and P.~Abbeel.
\newblock High-dimensional continuous control using generalized advantage
  estimation.
\newblock {\em CoRR}, abs/1506.02438, 2015.

\bibitem{PPO}
J.~Schulman, F.~Wolski, P.~Dhariwal, A.~Radford, and O.~Klimov.
\newblock Proximal policy optimization algorithms.
\newblock {\em CoRR}, abs/1707.06347, 2017.

\bibitem{ShelhamerMAD16Inverse}
E.~Shelhamer, P.~Mahmoudieh, M.~Argus, and T.~Darrell.
\newblock Loss is its own reward: Self-supervision for reinforcement learning.
\newblock {\em CoRR}, abs/1612.07307, 2016.

\bibitem{Siciliano2007}
B.~Siciliano and O.~Khatib.
\newblock {\em Springer Handbook of Robotics}.
\newblock Springer-Verlag, Berlin, Heidelberg, 2007.

\bibitem{Silberman2012}
N.~Silberman, D.~Hoiem, P.~Kohli, and R.~Fergus.
\newblock {\em Indoor Segmentation and Support Inference from RGBD Images},
  pages 746--760.
\newblock Springer Berlin Heidelberg, Berlin, Heidelberg, 2012.

\bibitem{Silver1140}
D.~Silver, T.~Hubert, J.~Schrittwieser, I.~Antonoglou, M.~Lai, A.~Guez,
  M.~Lanctot, L.~Sifre, D.~Kumaran, T.~Graepel, T.~Lillicrap, K.~Simonyan, and
  D.~Hassabis.
\newblock A general reinforcement learning algorithm that masters chess, shogi,
  and go through self-play.
\newblock {\em Science}, 362(6419):1140--1144, 2018.

\bibitem{singh2012unsupervised}
S.~Singh, A.~Gupta, and A.~A. Efros.
\newblock Unsupervised discovery of mid-level discriminative patches.
\newblock In A.~Fitzgibbon, S.~Lazebnik, P.~Perona, Y.~Sato, and C.~Schmid,
  editors, {\em Computer Vision -- ECCV 2012}, pages 73--86, Berlin,
  Heidelberg, 2012. Springer Berlin Heidelberg.

\bibitem{spelke2007core}
E.~S. Spelke and K.~D. Kinzler.
\newblock Core knowledge.
\newblock {\em Developmental science}, 10(1):89--96, 2007.

\bibitem{pmlr-v80-srinivas18b}
A.~Srinivas, A.~Jabri, P.~Abbeel, S.~Levine, and C.~Finn.
\newblock Universal planning networks: Learning generalizable representations
  for visuomotor control.
\newblock In J.~Dy and A.~Krause, editors, {\em Proceedings of the 35th
  International Conference on Machine Learning}, volume~80 of {\em Proceedings
  of Machine Learning Research}, pages 4732--4741, Stockholmsmässan, Stockholm
  Sweden, 10--15 Jul 2018. PMLR.

\bibitem{Sun16DeepCoral}
B.~Sun and K.~Saenko.
\newblock Deep {CORAL:} correlation alignment for deep domain adaptation.
\newblock {\em CoRR}, abs/1607.01719, 2016.

\bibitem{sutton}
R.~S. Sutton and A.~G. Barto.
\newblock {\em Introduction to Reinforcement Learning}.
\newblock MIT Press, Cambridge, MA, USA, 1st edition, 1998.

\bibitem{Szegedy2013adversarial}
C.~Szegedy, W.~Zaremba, I.~Sutskever, J.~Bruna, D.~Erhan, I.~J. Goodfellow, and
  R.~Fergus.
\newblock Intriguing properties of neural networks.
\newblock {\em CoRR}, abs/1312.6199, 2013.

\bibitem{Tsybakov2008nonparametric}
A.~B. Tsybakov.
\newblock {\em Introduction to Nonparametric Estimation}.
\newblock Springer Publishing Company, Incorporated, 1st edition, 2008.

\bibitem{Tzeng15adaptation}
E.~Tzeng, J.~Hoffman, T.~Darrell, and K.~Saenko.
\newblock Simultaneous deep transfer across domains and tasks.
\newblock {\em CoRR}, abs/1510.02192, 2015.

\bibitem{TzengHSD17Adversarial}
E.~Tzeng, J.~Hoffman, K.~Saenko, and T.~Darrell.
\newblock Adversarial discriminative domain adaptation.
\newblock {\em CoRR}, abs/1702.05464, 2017.

\bibitem{TzengHZSD14confusion}
E.~Tzeng, J.~Hoffman, N.~Zhang, K.~Saenko, and T.~Darrell.
\newblock Deep domain confusion: Maximizing for domain invariance.
\newblock {\em CoRR}, abs/1412.3474, 2014.

\bibitem{vandenOord2018predictive}
A.~van~den Oord, Y.~Li, and O.~Vinyals.
\newblock Representation learning with contrastive predictive coding.
\newblock {\em CoRR}, abs/1807.03748, 2018.

\bibitem{Vincent:2008:ECR:1390156.1390294}
P.~Vincent, H.~Larochelle, Y.~Bengio, and P.-A. Manzagol.
\newblock Extracting and composing robust features with denoising autoencoders.
\newblock In {\em Proceedings of the 25th International Conference on Machine
  Learning}, ICML '08, pages 1096--1103, New York, NY, USA, 2008. ACM.

\bibitem{Wang_2015_ICCV}
X.~Wang and A.~Gupta.
\newblock Unsupervised learning of visual representations using videos.
\newblock In {\em The IEEE International Conference on Computer Vision (ICCV)},
  December 2015.

\bibitem{acer}
Z.~Wang, V.~Bapst, N.~Heess, V.~Mnih, R.~Munos, K.~Kavukcuoglu, and
  N.~de~Freitas.
\newblock Sample efficient actor-critic with experience replay.
\newblock {\em CoRR}, abs/1611.01224, 2016.

\bibitem{xiazamirhe2018gibsonenv}
F.~Xia, A.~R.~Zamir, Z.-Y. He, A.~Sax, J.~Malik, and S.~Savarese.
\newblock Gibson env: real-world perception for embodied agents.
\newblock In {\em Computer Vision and Pattern Recognition (CVPR), 2018 IEEE
  Conference on}. IEEE, 2018.

\bibitem{Xiang2017pose}
Y.~Xiang, T.~Schmidt, V.~Narayanan, and D.~Fox.
\newblock Posecnn: {A} convolutional neural network for 6d object pose
  estimation in cluttered scenes.
\newblock {\em CoRR}, abs/1711.00199, 2017.

\bibitem{Yang2018ScenePriors}
W.~Yang, X.~Wang, A.~Farhadi, A.~Gupta, and R.~Mottaghi.
\newblock Visual semantic navigation using scene priors.
\newblock {\em CoRR}, abs/1810.06543, 2018.

\bibitem{yu2018imitation}
T.~Yu, P.~Abbeel, S.~Levine, and C.~Finn.
\newblock One-shot hierarchical imitation learning of compound visuomotor
  tasks.
\newblock {\em CoRR}, abs/1810.11043, 2018.

\bibitem{taskonomy2018}
A.~R. Zamir, A.~Sax, W.~B. Shen, L.~J. Guibas, J.~Malik, and S.~Savarese.
\newblock Taskonomy: Disentangling task transfer learning.
\newblock In {\em IEEE Conference on Computer Vision and Pattern Recognition
  (CVPR)}. IEEE, 2018.

\bibitem{zhang2016colorful}
R.~Zhang, P.~Isola, and A.~A. Efros.
\newblock Colorful image colorization.
\newblock In {\em European Conference on Computer Vision}, pages 649--666.
  Springer, 2016.

\bibitem{iss}
Y.~Zhong.
\newblock Intrinsic shape signatures: A shape descriptor for 3d object
  recognition.
\newblock In {\em 2009 IEEE 12th International Conference on Computer Vision
  Workshops, ICCV Workshops}, pages 689--696, Sept 2009.

\bibitem{MITplaces}
B.~Zhou, A.~Lapedriza, A.~Khosla, A.~Oliva, and A.~Torralba.
\newblock Places: A 10 million image database for scene recognition.
\newblock {\em IEEE Transactions on Pattern Analysis and Machine Intelligence},
  2017.

\bibitem{Zhu_2017_ICCV}
Y.~Zhu, D.~Gordon, E.~Kolve, D.~Fox, L.~Fei-Fei, A.~Gupta, R.~Mottaghi, and
  A.~Farhadi.
\newblock Visual semantic planning using deep successor representations.
\newblock In {\em The IEEE International Conference on Computer Vision (ICCV)},
  Oct 2017.

\end{thebibliography}
    }

    \end{document}

%%%%%%%%% TITLE
\title{Mid-Level Visual Representations Improve Generalization and Sample Efficiency for Learning Visuomotor Policies \\ \vspace{3mm}\large{Supplementary Material}}

\maketitle
\begin{abstract}
The following items are provided in the supplementary material:
 \renewcommand{\labelenumii}{\Roman{enumii}}
\begin{enumerate}[topsep=0pt,itemsep=-1ex,partopsep=1ex,parsep=1ex]
    \item A video of the included materials
    \item Clip of policies generalizing to real robot
    \item Frame-by-frame videos of feature networks in our environments
    \item Evaluation in additional test environments
    \item Universality results in VizDoom
    \item Baselines with 4.3M data frames
    \item Random Feature Set Baseline
    \item Numerical results for significance tests in the paper
    \item Plots of training and test performance for all features
    \item Raw reward curves
    \item Complete active task specification (observation space, action space, etc.)
    \item Description of mid-level vision tasks
    \item Full Description of Mid-Level Perception Module BIP
    \item Formal description of PPO with replay
    \item Significance tests with adjustments for cluster effects
    \item Experimental setup details (hyperparameters, architectures, train/test splits, etc.)
\end{enumerate}
\end{abstract}

% \section{Representation Selection Module Robustness}

% We examined the behavior of the feature selection module as we replace the options that the expert may choose from. Instead of the one set of well-performing features (Object Classification) we insert another option (Scene Classification). As shown in figure~\ref{fig:expert_ablation}, this leads the selection module to select the new best-performing feature (Scene Classification).

% \section{Insertion into Taskonomy}
\section{Policies on Real Robots}
Please see the included video for our example of policies generalizing to a real robot. The video shows our results from training a a curvature-based visual-target navigation policy on a simulated Turlebot2. We then then tested the trained policy (not fine-tuned) on a real Turtlebot2. The video shows that the robot is able to succesfully navigat to the correct cube, even in the presence of several distracting boxes which were not present during training.

\section{Frame-by-Frame Features}

The included video also includes videos that visualize the intermediate representations from our mid-level visual networks. The videos show frame-by-frame images from one of our testing environments, and the corresponding visualizations of the actual representations. We also provide a full list of studied features in Section~\ref{sec:task_dict}.
% This folder contains several videos, each of which shows the mid-level features evaluated on a different robotic agent. In Gibson the videos are for the Husky (default agent in this paper), Turtlebot, or roboschool Ant. In Doom, the agent is always the Doom marine.

\section{Evaluation in Multiple Test Environments}

% \begin{figure}
%     \centering
%     \includegraphics[width=\textwidth]{supmat_figures/expert_ablation.pdf}
%     \caption{\footnotesize{\textbf{Replacing features in selection module.} The agent selects the new best features (Scene Classification) when it is substituted for the old best feature (Object Classification). In each of the above panes the feature that eventually dominates is Scene Classification.}}
%     \label{fig:expert_ablation}
% \end{figure}
In order to rule out that our test environment was in some way anomalous, we repeated our testing in 9 additional environments. Figure~\ref{fig:multipletest} shows that the rewards in our main test are extremely highly correlated with the rewards in the alternate test environments. The feature rankings are extremely strongly correlated between the main test environment and the alternate test environments with a Spearman's rho of 0.93 (Navigation) and 0.85 (Exploration).

% enc of averaging
% We present test environment results (evaluated at the end of training) averaged over 10 test environments and all random seeds in Gibson for both the exploration and navigation tasks in the figure below. The results are presented in Figure~\ref{multipletestbarchart}. Our results agree with the single environment results presented in the main paper. In the navigation tasks, the semantic features perform significantly better than the baselines and both 2D and 3D geometric features. In the exploration tasks, the 3D geometric features outperform 2D geometric features and baselines, and usually show better performance than the semantic features.

\begin{figure}
    \centering
    \includegraphics[width=\columnwidth]{supmat_figures_iccv/navigate_moregibsonenvs.pdf}
    \includegraphics[width=\columnwidth]{supmat_figures_iccv/explore_moregibsonenvs.pdf}
    \caption{\footnotesize{\textbf{Correlation between main test environment and alternates.} The reward in the main test environment (x-axis) and the reward averaged over 9 other test environments (y-axis). Each point is the reward for a single feature (averaged over several seeds). Results are presented for both visual-target navigation (top) and visual exploration (bottom).}}
    \label{fig:multipletest}
\end{figure}

% \begin{figure}
%     \centering
%     \includegraphics[width=\columnwidth]{figures/multiple_test.pdf}
%     \label{multipletest}
%     \caption{\footnotesize{\textbf{Results in multiple Gibson testing environments.} Blue features are semantic, red features are 3D geometric, and green features are 2D geometric. Semantic features perform best in the navigation task, while 3D geometric features perform best in the exploration task.}}
%     \label{multipletestbarchart}
% \end{figure}

% \newpage

\section{Universality Experiments in VizDoom}

To address the possibility that our results are due to idiosyncrasies of the Gibson environment, we implemented navigation and exploration in Doom to evaluate whether one can expect to see similar effects hold across environments.  Figure~\ref{fig:doom_rank_reversal} shows that, like in Gibson, there is no universal feature. In fact, there seems to be an even larger tradeoff here. 

We again find the geometric/semantic distinction from Gibson appearing in Doom, and the results are highly statistically significant. Figure~\ref{fig:doom_rank_reversal_significance} shows rank reversal (in VizDoom) between the same tasks that we showed in Gibson. 

\begin{figure}
    \centering
    \includegraphics[width=\columnwidth]{supmat_figures_iccv/doom_gibson_feature_tradeoff.pdf}
    \caption{\footnotesize{\textbf{Rank reversal in visual tasks and no universal feature.} Scatterplots of the rank of feature performance on navigation (x-axis) and exploration (y-axis) in both Gibson (right) and Doom (left). The fact that there is no feature on the bottom left corner means there is no single universal feature|and the fact that there is a blank region there means there is also no \emph{almost} universal feature. The feature with the maximum F-score requires giving up 3-4 ranks for each task.}}    \label{fig:doom_rank_reversal}
\end{figure}

\begin{figure}
    \centering
    \includegraphics[width=\columnwidth]{figures/rank_reversal.pdf}
    \caption{\footnotesize{\textbf{Rank reversal significance graphs.} Arrows graphs indicate which features are better for a given downstream task. Heavier arrows indicate more significant results (lower $\alpha$-level). \textcolor{navigation}{Blue arrows} point toward tasks that are better for navigation and \textcolor{exploration}{red arrows} point toward tasks that better support exploration. Lack of an arrow indicates the performance difference was not statistically significant. That there is no node with all incoming arrows demonstrates the lack of a universal feature. The essentially complete bipartite structure in the Gibson graph shows that navigation is characteristically semantic while exploration is geometric.}}    \label{fig:doom_rank_reversal_significance}
\end{figure}

\subsection{When Features May NOT Help}

We also found that features were more robust to changes in texture than learning from scratch. While \emph{scratch} achieves the highest final performance when the agent learns in a video game environment where there are many train textures that emulate the test textures, \emph{scratch} fails to generalize when there is little or no variation in texture during training. On the other hand, feature-based agents are able to generalize even without texture randomization, as shown in fig.~\ref{fig:doom_texture_generalization}. This suggests reinforcement learning works well in the limit of infinite data, and when the test distribution is well approximated by the training distribution. Our findings suggest a crucial subtlety: that when with limited data and when the training distribution is not wholly representative of the test distribution, some inductive bias could be helpful. 

\begin{figure}
    \includegraphics[width=\columnwidth]{figures/doom_generalization_curves_rel_reward.pdf}
    \caption{\footnotesize{\textbf{Features generalize to new axes of variation.} In ViZDoom, feature-based agents (right two columns) generalize to new textures even when not exposed to texture variation in training (top row), while agents trained from scratch suffer a significant drop in performance (left, top).}}
    \label{fig:doom_texture_generalization}
\end{figure}

\section{Baselines with 4.3M frames}

All our agents were trained on the same number of environment frames. However, the encoders were trained using some static images of indoor environments. While these images are not accessible to the usual RL setup, they are additional data that is available to the feature-based agent. These encoders were trained on 4 million images of indoor environments. Figures~\ref{fig:long_train_nav}, \ref{fig:long_train_explore}, and \ref{fig:long_train_planning} show that, even after training our \emph{tabula rasa} agents for an additional 4 million frames, mid-level-based agents still achieve superior performance. Those figures also show the curves when we train the feature-based agents for longer|just to show what those curves look like. Autoencoder still does not outperform \emph{scratch} in the test environment.

\begin{figure}
    \includegraphics[width=\columnwidth]{supmat_figures_iccv/long_training_navigation.pdf}
    \caption{\footnotesize{\textbf{Long-training Navigation.} Navigation policies trained for 4.3 milion frames, to compare our policies in the main paper to \emph{data equivalent scratch} policies (scratch policies that have seen the same number of frames as the feature encoders \emph{and} the mid-level policies combined.}}
    \label{fig:long_train_nav}
\end{figure}
    
\begin{figure}
    \includegraphics[width=\columnwidth]{supmat_figures_iccv/long_training_exploration.pdf}
    \caption{\footnotesize{\textbf{Long-training Exploration.} Exploration policies trained for 4.3 million frames, as above.}}
    \label{fig:long_train_explore}
\end{figure}

\begin{figure}
    \centering
    \includegraphics[width=0.45\columnwidth]{supmat_figures_iccv/long_training_planning.pdf}
    \caption{\footnotesize{\textbf{Long-training Planning.} Local planning policies trained for 4.3 million frames, as above.}}
    \label{fig:long_train_planning}
\end{figure}

\section{Random Feature Set Baselines}

How useful is the feature set proposed by the perception module? Will any feature set work? We randomly uniformly selected five feature sets and evaluated their performance on our active tasks.  Table~\ref{fig:random_feature_sets} shows that the solver-suggested feature set performs much better than the randomly selected sets on navigation, and comparably on the other two tasks. We hypothesize that the high performance of random features on exploration is due to a larger number of geometric-based tasks in our task dictionary, which tend to excel at the exploration task, while the relatively worse performance on navigation is due to the small number of semantic features in our dictionary. Our perception module ensures coverage of both kinds of features, which leads to good performance on both navigation and exploration.\\

It is notable that both our perception module and random sets of features outperform \emph{tabula rasa} learning (and our other baselines) by a wide margin. 

%\begin{figure}
%    \centering
%    \includegraphics[width=0.45\columnwidth]{supmat_figures_iccv/random_feature_sets.png} \caption{\footnotesize{\textbf{Task performance using random feature sets.} \todo{update}}}
%    \label{fig:random_feature_sets}
%\end{figure}

\begin{table}[]
\begin{tabular}{lllllll}
\hline
\multirow{2}{*}{Features} & \multicolumn{2}{l}{Navigation} & \multicolumn{2}{l}{Exploration} & \multicolumn{2}{l}{Planning} \\ \cline{2-7} 
                          & \textit{Ours}      & Rand.     & \textit{Ours}   & Rand.         & \textit{Ours}     & Rand.    \\ \hline
2                         & \textbf{1.7}       & 1.5       & 1.2             & \textbf{1.4}  & 1.2               & 1.2      \\ \hline
3                         & \textbf{2.1}       & 1.8       & 1.2             & \textbf{1.3}  & 1.2               & 1.2      \\ \hline
4                         & \textbf{2.4}       & 1.9       & \textbf{1.4}    & 1.3           & 1.2               & 1.2      \\ \hline
\end{tabular}
\caption{Our perception module beats random sets of features on navigation by a wide margin and performs just as well on exploration and planning.}
\label{fig:random_feature_sets}
\end{table}

\section{Numerical Results for Significance Analysis}

In Figure~\ref{fig:h2_numbers} we present the exact p-values for the \emph{hypothesis II} experiments in the main paper.

\begin{figure*}[ht]
  % NAV
  \hspace{-5mm}
  \begin{minipage}{0.3\textwidth}
      \centering
      \textbf{Navigation}\\
      \vspace{-2mm}     \noindent\rule{5cm}{0.4pt}\\
      \vspace{1mm}
      \begin{minipage}{0.47\linewidth}
        \hspace{-5mm}
        \includegraphics[width=\textwidth]{figures/gibson_navigate_select.pdf}
      \end{minipage}
      \hspace{-6mm}
      \begin{minipage}{0.5\linewidth}
          \centering
          \footnotesize
          \setlength\tabcolsep{2pt} % default value: 6pt
          \begin{tabular}{ c|c|c| } 
            \cline{2-3}
            \textbf{Feature} & \textbf{r} & \textbf{p-val} \\
             \hline
             Obj. Cls.            &  \tiny{5.91}  &  .001 \\
             Sem. Segm.           &  \tiny{5.87}  &  .001 \\
             Curvature            &  \tiny{4.75}  &  .002 \\
             Scene Cls.           &  \tiny{3.07}  &  .003 \\
             2.5D Segm.           &  \tiny{3.01}  &  .002 \\
             2D Segm.             &  \tiny{1.99}  &  .003 \\
             Distance             &  \tiny{1.74}  &  .003 \\
             Occ. Edges           &  \tiny{.38}  &  .009 \\
             Vanish. Pts.         &  \tiny{.39}  &  .019 \\
             Reshading            &  \tiny{.21}  &  .021 \\
             2D Edges             &  \tiny{.12}  &  .006 \\
             Normals              &  \tiny{-.50}  &  .035 \\
             Jigsaw               &  \tiny{-.86}  &  .122 \\
             3D Keypts.           &  \tiny{-1.08}  &  .112 \\
             Layout               &  \tiny{-1.14}  &  .057 \\
             Autoenc.             &  \tiny{-1.16}  &  .043 \\
             Rand. Proj.          &  \tiny{-2.12}  &  .083 \\
            \rowcolor{gray!50} Blind                &  \tiny{-3.20}  &  .755 \\
            \rowcolor{gray!50} Pix-as-state         &  \tiny{-4.30}  &  .856 \\
            \rowcolor{gray!50} 2D Keypts.           &  \tiny{-6.10}  &  .922 \\
            \rowcolor{gray!50} In-painting          &  \tiny{-6.57}  &  .971 \\
            \rowcolor{gray!50} Denoising            &  \tiny{-6.47}  &  .981 \\
            \cline{2-3}
          \end{tabular}
      \end{minipage}
  \end{minipage}
  \begin{minipage}{0.3\textwidth}
      \centering
      \textbf{Exploration}\\
      \vspace{-2mm}     \noindent\rule{5cm}{0.4pt}\\
      \vspace{1mm}
      \begin{minipage}{0.48\linewidth}
        \centering
        \hspace{-2mm}
        \includegraphics[width=\textwidth]{figures/gibson_explore_select.pdf}
      \end{minipage}
      \hspace{-2mm}
      \begin{minipage}{0.5\linewidth}
          \centering
          \footnotesize
          \setlength\tabcolsep{2pt} % default value: 6pt
          \begin{tabular}{ c|c|c| } 
            \cline{2-3}
            \textbf{Feature} & \textbf{r} & \textbf{p-val} \\
            \hline
             Distance             &  \tiny{5.90}  &  .015 \\
             Reshading            &  \tiny{5.79}  &  .003 \\
             3D Keypts.           &  \tiny{5.27}  &  .004 \\
             Curvature            &  \tiny{5.12}  &  .027 \\
            \rowcolor{gray!50} 2.5D Segm.           &  \tiny{5.60}  &  .056 \\
            \rowcolor{gray!50} Layout               &  \tiny{4.78}  &  .108 \\
            \rowcolor{gray!50} 2D Edges             &  \tiny{4.87}  &  .120 \\
            \rowcolor{gray!50} Normals              &  \tiny{5.26}  &  .143 \\
            \rowcolor{gray!50} Scene Cls.           &  \tiny{4.67}  &  .152 \\
            \rowcolor{gray!50} Obj. Cls.            &  \tiny{4.80}  &  .187 \\
            \rowcolor{gray!50} 2D Segm.             &  \tiny{4.47}  &  .406 \\
            \rowcolor{gray!50} Jigsaw               &  \tiny{4.47}  &  .455 \\
            \rowcolor{gray!50} Rand. Proj.          &  \tiny{4.33}  &  .500 \\
            \rowcolor{gray!50} Vanish. Pts.         &  \tiny{4.24}  &  .500 \\
            \rowcolor{gray!50} Pix-as-state         &  \tiny{4.20}  &  .531 \\
            \rowcolor{gray!50} Blind                &  \tiny{4.21}  &  .545 \\
            \rowcolor{gray!50} 2D Keypts.           &  \tiny{4.21}  &  .682 \\
            \rowcolor{gray!50} In-painting          &  \tiny{4.30}  &  .697 \\
            \rowcolor{gray!50} Autoenc.             &  \tiny{4.11}  &  .815 \\
            \rowcolor{gray!50} Sem. Segm.           &  \tiny{3.67}  &  .857 \\
            \rowcolor{gray!50} Occ. Edges           &  \tiny{3.85}  &  .864 \\
            \rowcolor{gray!50} Denoising            &  \tiny{3.59}  &  .962 \\
            \cline{2-3}
          \end{tabular}
      \end{minipage}
  \end{minipage}
  \hspace{1mm}
  \begin{minipage}{0.3\textwidth}
      \centering
      \textbf{Local Planning}\\
      \vspace{-2mm}     \noindent\rule{5cm}{0.4pt}\\
      \vspace{1mm}
      \begin{minipage}{0.48\linewidth}
        \includegraphics[width=\textwidth]{figures/gibson_planning_select.pdf}
      \end{minipage}
      \hspace{-2mm}
      \begin{minipage}{0.5\linewidth}
          \centering
          \footnotesize
          \setlength\tabcolsep{2pt} % default value: 6pt
          \begin{tabular}{ c|c|c| } 
            \cline{2-3}
            \textbf{Feature} & \textbf{r} & \textbf{p-val} \\
             \hline
             3D Keypts.           &  \tiny{15.45}  &  .015 \\
             Normals              &  \tiny{15.10}  &  .000 \\
             Curvature            &  \tiny{14.84}  &  .003 \\
             Distance             &  \tiny{14.56}  &  .001 \\
             2.5D Segm.           &  \tiny{14.50}  &  .001 \\
             Sem. Segm.           &  \tiny{14.49}  &  .000 \\
             Scene Cls.           &  \tiny{14.20}  &  .001 \\
             Occ. Edges           &  \tiny{14.20}  &  .001 \\
             Reshading            &  \tiny{14.12}  &  .000 \\
             Layout               &  \tiny{14.12}  &  .015 \\
             Obj. Cls.            &  \tiny{13.95}  &  .000 \\
             2D Segm.             &  \tiny{13.86}  &  .001 \\
             Denoising            &  \tiny{13.54}  &  .000 \\
             In-painting          &  \tiny{13.28}  &  .000 \\
             Jigsaw               &  \tiny{13.17}  &  .012 \\
             2D Edges             &  \tiny{13.16}  &  .008 \\
             Vanish. Pts.         &  \tiny{12.14}  &  .028 \\
             2D Keypts.           &  \tiny{11.99}  &  .050 \\
             Autoenc.             &  \tiny{11.39}  &  .155 \\
            \rowcolor{gray!50} Pix-as-state         &  \tiny{10.22}  &  .654 \\
            \rowcolor{gray!50} Rand. Proj.          &  \tiny{8.93}  &  .892 \\
            \rowcolor{gray!50} Blind                &  \tiny{9.83}  &  .929 \\
            \cline{2-3}
          \end{tabular}
      \end{minipage}
  \end{minipage}
  \vspace{0.5mm}
  \caption{\footnotesize{\textbf{Features vs. scratch.} The plots above show training and test performance of scratch vs. some selected features throughout training. For all tasks there is a significant gap between train/test performance for scratch, and a much smaller one for the best feature. The tables above show significance tests of the performance of feature-based agents vs from scratch in Gibson. P-values come from a Wilcoxon rank sum test, adjusted for multiple hypothesis testing with a FDR of 20\%. Significant rows are in white, and these blocks are ordered by average episode reward. The plots show training and testing curves, and there are significant gaps between them. Scratch often fails to generalize (bottom), while feature-based agents generalize better (top). Sometimes, models may appear to learn in the training environment, but they fail at test time|underscoring the importance of a good test environment in RL.}}
  \label{fig:h2_numbers}
\end{figure*}

% \cleardoublepage

\section{Mid-Level Perception Module with Feature Interactions}

The Boolean Integer Program (BIP) described in the main text works|but does not account for interactions between features. For example, what if the combination of two features is much stronger than either feature separately? This section details a variant of the main BIP that handles these interactions. It selects a set of transfers which \emph{may have one or more sources}.

The set of all transfers (edges), $E$, is indexed by $i$ and each edge has the form $(\{s^i_1, \dotso, s^i_{m_i}\}, t^i)$ (for $\{s^i_1, \dotso ,s^i_{m_i}\} \subset \mathcal{S}$ and $t^i \in \mathcal{T}$). We shall also index the target features $\mathcal{\Phi}$ by $j$ so that in this section, $i$ indexes edges and $j$ indexes target features. As in \cite{taskonomy2018}, we define operators returning target and sources of an edge:

%	\vspace{-2mm}
% \begin{gather*}
\begin{align*}
\big(\{s^i_1, \dotso, s^i_{m_i}\}, t^i\big) \hspace{2mm} & \xmapsto[]{sources} \hspace{1mm} \{s^i_1, \dotso, s^i_{m_i}\}\\
\big(\{s^i_1, \dotso, s^i_{m_i}\}, t^i\big) \hspace{2mm} & \xmapsto[]{target} \hspace{4mm} t^i.
\end{align*}
% \end{gather*}
%We define analogous functions on indices $i$. 

We encode selecting a feature $t$ to include in the set as including the transfer $\big(\{t\}, t\big)$. 
%The solver then chooses which transfers to use, given the following parameters: 

The arguments of the problem are
\begin{enumerate}
    \item $\delta$, a given maximum covering distance
    \item $p_i$, a measure of performance on a target from each of its transfers (i.e. the affinities from~\cite{taskonomy2018})
    \item $x \in \mathbb{R}^{|E| + |\Phi|}$, a boolean variable indicating whether each transfer and each feature in our set
\end{enumerate}

The BIP is parameterized by a vector $x$ where each transfer and each feature in our set is represented by a binary variable; $x$ indicates which nodes are selected for the covering set and a satisfying minimum-distance transfer from each unselected feature to one in the set. The canonical form for a BIP is: 
% \vspace{-3mm}
\begin{align*}
\text{minimize: } & \mathbbm{1}^T x\,,\\
\text{subject to: } & Ax \preceq b \text{ and } x \in \{0, 1\}^{|E| + |\Phi|}\,.
\end{align*}

Each element $c_i$ for a transfer is the product of the importance of its target task and its transfer performance:
%\vspace{-3mm}
\begin{equation}
c_i := r_{\textit{target}(i)} \cdot p_i\,.
\end{equation}
Hence, the \emph{collective} performance on all targets is the summation of their individual AHP performance, $p_i$, weighted by the user specified importance, $r_i$.

Now we add three types of constraints via matrix $A$ to enforce each feasible solution of the BIP instance corresponds to a valid subgraph for our transfer learning problem: \emph{Constraint I:} no feature is too far away from the covering set, \emph{Constraint II:} if a transfer is included in the subgraph, all of its source nodes/tasks must be included too, \emph{Constraint III:}  each target task has exactly one transfer in. 

\emph{Constraint I: No feature is too far}. We ensure that this by disallowing edges whose ``distance'' is too large. In our case, we use affinities (so that higher values indicate a better transfer). Therefore, we filter out edges with small affinities by redefining $$E \triangleq \{~e \in E~|~\text{affinity}(e) \leq \delta~\}.$$

\emph{Constraint II: All necessary sources are present}.
For each row $a_i$ in $A$ we require $a_i \cdot x \leq b_i$, where 
%\vspace{1mm}
\begin{align}
a_{i,k} &\defeq \begin{cases}
|\textit{sources}(i)| & \text{if } k = i\\
-1 & \text{if } (k - |E|) \in \textit{sources}(i)\\
0 & \text{otherwise}\\
\end{cases}\\
b_{i} &= 0.
\end{align}
%	\vspace*{-\baselineskip}

\emph{Constraint II: One transfer per feature}.
Via two rows $a_{|E| + j}$ and $a_{|E| + j + 1}$, we enforce that target $j$ has exactly one transfer:
% \vspace{-3mm}
\begin{equation}
\begin{array}{l}
    a_{|E| + j, i} \hspace{10.2pt} \defeq \hspace{7.3pt} \mathbbm{1}_{\{\textit{target}(i)=j\}},\hspace{0.5cm} b_{|E| + j} \hspace{10.2pt} \defeq +1, \\ 
    a_{|E| + j+1, i} \defeq -\mathbbm{1}_{\{\textit{target}(i)=j\}},\hspace{0.5cm} b_{|E| + j+1} \defeq -1.
\end{array}
\end{equation}

Those elements of A not explicitly defined above are set to 0. The problem is now a valid BIP and can be optimally solved in a fraction of a second~\cite{gurobi}.

The above program finds a (not necessarily unique) minimum-size covering set with a covering distance at most $\delta$. Given that our feature set has only a finite number of distances, we can find the smallest $\delta$ by solving a sequence of these BIPs (e.g. with binary search).

\section{PPO with Experience Replay}

In this section we describe our off-policy PPO variant which decorrelates the samples within each batch and provides more stable, sample-efficient learning for the Gibson environment with only one rollout worker. 

\subsection{Off-Policy Policy Gradient}

In the most general policy gradient setup, we attempt to maximize the objective function:

\begin{equation}
    \mathbb{E}_{\tau \sim \pi_{\theta}(\tau)} \Big[ \log \pi_{\theta}(a_t | s_t) \hat{A}_t \Big]
\end{equation}

where $\hat{A}_t$ is the advantage function at timestep $t$ (some sufficient statistic for the value of a policy at timestep $t$, in our experiments we choose the generalized advantage estimator \cite{gae}) and $\tau$ is a trajectory drawn from the current policy. The right side of the equation is our estimate of the objective using data sampled from the environment under the policy. Using importance sampling, we can approximate this objective by sampling from a different distribution over trajectories, and instead optimize:

\begin{equation}
    \mathbb{E}_{\tau \sim \pi_{\theta_{\text{old}}}(\tau)} \Big[ \frac{\pi_{\theta}(a_t | s_t)}{\pi_{\theta_{\text{old}}}(a_t | s_t)} \hat{A}_t \Big]
\end{equation}

\subsection{Off-Policy PPO Clipping}

In practice, if we were to directly optimize the objective in equation (2), this would lead to dramatically unstable gradient updates due to both the high variance of both the importance sampling ratio and $\hat{A}_t$ (in our actor-critic framework, we are learning $\hat{A}$ as the critic as training progresses, so in the beginning of training $\hat{A}$ is especially unstable).

Proximal policy optimization decreases variance by clipping the policy ratio, minimizing the surrogate objective:

\begin{equation}
   \mathbb{E}_{\tau \sim \pi_{\theta_{\text{old}}}(\tau)} \min\Big[ r_t(\theta) \hat{A}_t, \text{clip}(1 - \epsilon, 1+\epsilon, r_t(\theta)) \hat{A}_t \Big]
\end{equation}

\begin{equation}
r_t(\theta) = \frac{\pi_{\theta}(a_t | s_t)}{\pi_{\theta_{\text{old}}} (a_t | s_t)}
\end{equation}

 This objective acts as a first-order trust region, preventing large policy updates that drastically change the policy. For detailed theoretical justification of this objective, see the PPO paper \cite{PPO}. In the original PPO paper, $\pi_{\theta_\text{old}}$ is only the distribution of on-policy trajectories from parallel workers. Since we are constrained to only one worker, sampling from on-policy trajectories only would lead to batches with highly correlated samples, leading to overfitting. Instead, we maintain a replay buffer, and draw multiple trajectories from the replay buffer at every update, treating the policy at the time when the trajectories were executed as $\pi_{\theta_\text{old}}$. To calculate the advantage, we reevaluate the advantage function during the update using the current critic on the states on the sampled trajectories from the replay buffer. The number of on-policy and off-policy trajectories that we sample under this formulation is a hyperparameter which we tune and report.

\section{Active Task Dictionary}

\subsection{Reinforcement Learning State}

In both tasks, the agent's state is simply a stream of RGB images. In our case, the agent receives the 4 most recent frames. The agent receives no proprioception information such as its momentum or joint locations, GPS, IMU, etc. For exploration, the agent's state also includes the egocentric occupancy map as a bitmap image. All actions are available to the agent at all times, although the physical constraints of the space are implicitly enforced through the physics engines (i.e., if the agent goes forward into a wall, the agent will bounce back off the wall). 

\subsection{Action Space}

We assume a low-level controller for robot actuation, enabling a high-level action space of\vspace{-2mm} $$\mathcal{A}=\small{\{}\texttt{\small{turn\_left}},~\texttt{\small{turn\_right}},~ \texttt{\small{move\_forward}}\small{\}}\vspace{-2mm}.$$. The forward action is a translation in the direction of the robot's heading of $0.1$ m, and the turn actions represent in-place rotations of $\pm 0.24$ radians (about 14 degrees).

\subsection{Visual-Target Navigation}

The reward structure for navigation is sparse, reflecting the conditions of the real-world where reward shaping for a new task is often difficult and can produce undesired behavior. \\

In Gibson, the agent has to navigate to a wooden crate with only an RGB camera frame in its observation space. We stack the 4 most recent frames and use that as the RL state. The agent gets a reward of $+10$ for hitting the target and a reward of $-0.025$ at each timestep for living. Hitting the target or going outside the predefined boundary (e.g., going to a different floor) terminates the episode. In VizDoom, the agent must navigate to a green torch. The agent gets a reward of $+100$ for hitting the target and a reward of $-1$ for living. 

\subsection{Visual Exploration}

The visual exploration task is designed to train an agent to use its RGB image to scan as much of its environment as possible in a limited amount of time. In both Gibson and VizDoom, the agent's goal is to uncover as many cells as possible using its myopic laser scanner. The cells uncovered at each timestep are counted by forming a point cloud using a narrow strip of the depth image from the midpoint of the image to the bottom, and then projecting the point cloud onto the ground plane. The ground plane is discretized into 1 m by 1 m cells. As observation, the agent receives this occupancy grid as a bitmap image, and all non-blind agents also receive the RGB camera frame, as above. In Gibson, the agent receives a reward of $+0.1$ for each cell unlocked.  In VizDoom, the agent receives a reward of $+1$ for each cell unlocked.

\subsection{Local Planning}

In local planning, the agent must navigate to a nonvisual target specified by coordinates in its own inertial reference frame. Its reward is proportional to the dense progress ($0.1 * (x_{t - 1}- x_{t})$), plus a dense reward for hitting the goal ($+20$), a penalty for living ($-0.05$), and an obstacle collision penalty ($-0.25$), which reflects desirable behavior of a \emph{local} planner, which should skillfully maneuver its environment while taking the most efficient path to the goal. The goal is sampled from a Gaussian distribution, $~\mathcal{N}(\mu=5~\text{meters}, \sigma^2=2~\text{m)}$.\\

Its observation state is a 3-vector $[r, \cos{\theta}, \sin{theta}]$ in the agent frame specifying the target vector, as well as the RGB camera images, as above.

\section{Mid-Level Task Dictionary}\label{sec:task_dict}

We present the tasks in alphabetical order in Figure~\ref{tasks}. A detailed description of each task can be found the Taskonomy~\cite{taskonomy2018} supplementary material (Section 14). 

\begin{figure}[ht]
  \begin{minipage}{\columnwidth}
  \centering
  \small
  \begin{tabular}{ |c| } 
    \hline
    \textbf{Mid-Level Vision Task Dictionary ($\Phi$)} \\
    \hline
    \hline
    Autoencoding \\
    Classification, Semantic (1000-class) \\
    Classification, Scene \\
    Context Encoding (In-painting) \\
    Content Prediction (Jigsaw) \\
    Depth Estimation, Euclidean \\
    Edge Detection, 2D \\
    Edge Detection, 3D \\
    Keypoint Detection, 2D \\
    Keypoint Detection, 3D \\
    Reshading \\
    Room Layout Estimation \\
    Segmentation, Unsupervised 2D \\
    Segmentation, Unsupervised 2.5D \\
    Segmentation, Semantic \\
    Surface Normal Estimation \\
    Vanishing Point Estimation \\
    \hline 
    
  \end{tabular}
  \end{minipage}
  \vspace{0.5mm}
  \caption{\footnotesize{\textbf{Feature Bank} The features we used for all experiments. See Taskonomy~\cite{taskonomy2018} suppplementary for more details.}}
  \label{tasks}
\end{figure}

\section{Full Train/Test curves}\label{sec:train_test_curves}

We present the training and test curves from our experiments for both tasks, for all features and baselines, and all random seeds. See ~\ref{navresults} and ~\ref{exploreresults}.

\cleardoublepage

\begin{figure*}[ht]
    \centering
    \vspace{-3mm}
    \textbf{Gibson Visual-Target Navigation Train/Test}\\
    \includegraphics[width=\textwidth]{figures/full_navigate.pdf}
    \vspace{-8mm}
    \caption{\footnotesize{\textbf{Full Learning Curves (Gibson Navigation).} Full train and test curves for all features and baselines in the navigation task in the Gibson environment.}}
    \label{navresults}
\end{figure*}

\begin{figure*}[ht]
    \centering
    \vspace{-3mm}
    \textbf{Gibson Visual Exploration Train/Test}\\
    \includegraphics[width=\textwidth]{figures/full_explore.pdf}
    \vspace{-8mm}
    \caption{\footnotesize{\textbf{Full Learning Curves (Gibson Exploration).} Full train and test curves for all features and baselines in the exploration task in the Gibson environment.}}
    \label{exploreresults}
\end{figure*}

\begin{figure*}[ht]
    \centering
    \vspace{-3mm}
    \textbf{Gibson Local Planning Train/Test}\\
    \includegraphics[width=\textwidth]{supmat_figures_iccv/gibson_coordinate_planning_raw.pdf}
    \vspace{-8mm}
    \caption{\footnotesize{\textbf{Full Learning Curves (Gibson Local Planning).} Full train and test curves for all features and baselines in the planning task in the Gibson environment.}}
    \label{exploreresults}
\end{figure*}

\begin{figure*}[ht]
    \centering
    \vspace{-3mm}
    \textbf{Gibson Visual-Target Navigation Raw Reward}\\
    \includegraphics[width=\textwidth]{figures/full_navigate.pdf}
    \vspace{-8mm}
    \caption{\footnotesize{\textbf{Full Learning Curves (Gibson Navigation).} Full train and test curves for all features and baselines in the navigation task in the Gibson environment.}}
    \label{navresults}
\end{figure*}

\begin{figure*}[ht]
    \centering
    \vspace{-3mm}
    \textbf{Gibson Visual Exploration Raw Reward}\\
    \includegraphics[width=\textwidth]{figures/full_explore.pdf}
    \vspace{-8mm}
    \caption{\footnotesize{\textbf{Full Learning Curves (Gibson Exploration).} Full train and test curves for all features and baselines in the exploration task in the Gibson environment.}}
    \label{exploreresults}
\end{figure*}

\begin{figure*}[ht]
    \centering
    \vspace{-3mm}
    \textbf{Gibson Local Planning Raw Reward}\\
    \includegraphics[width=\textwidth]{supmat_figures_iccv/gibson_coordinate_planning_raw.pdf}
    \vspace{-8mm}
    \caption{\footnotesize{\textbf{Full Learning Curves (Gibson Local Planning).} Full train and test curves for all features and baselines in the local planning task in the Gibson environment.}}
    \label{exploreresults}
\end{figure*}

\begin{figure*}[ht]
    \centering
    \vspace{-3mm}
    \textbf{Gibson Local Planning Raw Reward}\\
    \includegraphics[width=\textwidth]{figures/full_explore.pdf}
    \vspace{-8mm}
    \caption{\footnotesize{\textbf{Full Learning Curves (Gibson Local Planning).} Full train and test curves for all features and baselines in the planning task in the Gibson environment.}}
    \label{exploreresults}
\end{figure*}

\begin{figure*}[ht]
    \centering
    \vspace{-3mm}
    \textbf{Doom Visual-Target Navigation Raw Reward}\\
    \includegraphics[width=\textwidth]{supmat_figures/doom_navigate_plots.pdf}
    \vspace{-8mm}
    \caption{\footnotesize{\textbf{Full Learning Curves (Doom Navigation).} Full train and test curves for all features and baselines in the navigation task in the ViZDoom environment.}}
    \label{doomnavigationresults}
\end{figure*}

\begin{figure*}[ht]
    \centering
    \vspace{-3mm}
    \textbf{Doom Visual Exploration Raw Reward}\\
    \includegraphics[width=\textwidth]{supmat_figures/doom_explore_plots.pdf}
    \vspace{-8mm}
    \caption{\footnotesize{\textbf{Full Learning Curves (Doom Exploration).} Full train and test curves for all features and baselines in the exploration task in the ViZDoom environment.}}
    \label{doomexploreresults}
\end{figure*}

\cleardoublepage

\section{Cluster-based Wilcoxon Analysis}

In this section we present results using the cluster-aware Wilcoxon rank-sum test in the \texttt{clusrank}~\cite{clusrank} package in R. 
%We include the code in the file \texttt{significance\_testing.r}.

\subsection{Hypothesis II Experiments}

% \todo{go over this}
Figure~\ref{fig:h2_cluster_analysis} shows features that are significantly different from scratch, after adjusting for cluster effects (finite sample-estimation of seed performance).

\begin{figure*}[ht]
%   \hspace{-mm}
    \begin{minipage}{0.3\columnwidth}
    \begin{center}
      \textbf{Visual-Target Navigation}
      \footnotesize
      \begin{tabular}{ |c|c|c|c| } 
        \hline
        \textbf{Feature} & \textbf{Rew.} & \textbf{p-val.} & $\mathbf{i/m}$ \\
        \hline
        \hline
        Sem. Segm.           &  5.866  &  0.001  &  0.009 \\
        Obj. Cls.            &  5.905  &  0.001  &  0.018 \\
        Curvature            &  4.748  &  0.002  &  0.027 \\
        2.5D Segm.           &  3.012  &  0.002  &  0.036 \\
        Scene Cls.           &  3.067  &  0.003  &  0.045 \\
        2D Segm.             &  1.992  &  0.003  &  0.055 \\
        Distance             &  1.742  &  0.003  &  0.064 \\
        2D Edges             &  0.121  &  0.006  &  0.073 \\
        Occ. Edges           &  0.382  &  0.009  &  0.082 \\
        Vanish. Pts.         &  0.389  &  0.019  &  0.091 \\
        Reshading            &  0.206  &  0.021  &  0.100 \\
        Normals              &  -0.502  &  0.035  &  0.109 \\
        Autoenc.             &  -1.159  &  0.043  &  0.118 \\
        Layout               &  -1.138  &  0.057  &  0.127 \\
        Rand. Proj.          &  -2.124  &  0.083  &  0.136 \\
        3D Keypts.           &  -1.082  &  0.112  &  0.145 \\
        Jigsaw               &  -0.865  &  0.122  &  0.155 \\
        \hline
        Blind                &  -3.204  &  0.755  &  0.164 \\
        Pix-as-state         &  -4.297  &  0.856  &  0.173 \\
        2D Keypts.           &  -6.095  &  0.922  &  0.182 \\
        In-painting          &  -6.571  &  0.971  &  0.191 \\
        Denoising            &  -6.469  &  0.981  &  0.200 \\
        \hline
      \end{tabular}
    \end{center}
    \end{minipage}
    \begin{minipage}{0.3\columnwidth}
    \begin{center}
        \textbf{Visual Exploration}
        \footnotesize
        \begin{tabular}{ |c|c|c|c| } 
            \hline
            \textbf{Feature} & \textbf{Rew.} & \textbf{p-val.} &  $\mathbf{i/m}$ \\
            \hline
            \hline
            2.5D Segm.           &  5.897  &  0.011  &  0.009 \\
            Reshading            &  5.738  &  0.014  &  0.018 \\
            Distance             &  5.839  &  0.025  &  0.027 \\
            Layout               &  5.136  &  0.038  &  0.036 \\
            Curvature            &  5.091  &  0.038  &  0.045 \\
            3D Keypts.           &  4.939  &  0.053  &  0.055 \\
            \hline
            Normals              &  4.968  &  0.163  &  0.064 \\
            2D Edges             &  4.632  &  0.201  &  0.073 \\
            Scene Cls.           &  4.556  &  0.216  &  0.082 \\
            Obj. Cls.            &  4.699  &  0.236  &  0.091 \\
            2D Keypts.           &  4.440  &  0.337  &  0.100 \\
            Jigsaw               &  4.317  &  0.539  &  0.109 \\
            Blind                &  4.191  &  0.568  &  0.118 \\
            Autoenc.             &  4.113  &  0.577  &  0.127 \\
            Denoising            &  4.036  &  0.629  &  0.136 \\
            In-painting          &  4.061  &  0.693  &  0.145 \\
            2D Segm.             &  4.028  &  0.707  &  0.155 \\
            Pix-as-state         &  3.945  &  0.729  &  0.164 \\
            Rand. Proj.          &  3.782  &  0.754  &  0.173 \\
            Sem. Segm.           &  3.844  &  0.774  &  0.182 \\
            Vanish. Pts.         &  3.817  &  0.812  &  0.191 \\
            Occ. Edges           &  3.772  &  0.888  &  0.200 \\
            \hline
        \end{tabular}
    \end{center}
    \end{minipage}
    \begin{minipage}{0.30\columnwidth}
    \begin{center}
        \textbf{Local Planning}
        \footnotesize
        \begin{tabular}{ |c|c|c|c| } 
            \hline
            \textbf{Feature} & \textbf{Rew.} & \textbf{p-val.} &  $\mathbf{i/m}$ \\
            \hline
            \hline
            2.5D Segm.           &  5.897  &  0.011  &  0.009 \\
            Reshading            &  5.738  &  0.014  &  0.018 \\
            Distance             &  5.839  &  0.025  &  0.027 \\
            Layout               &  5.136  &  0.038  &  0.036 \\
            Curvature            &  5.091  &  0.038  &  0.045 \\
            3D Keypts.           &  4.939  &  0.053  &  0.055 \\
            \hline
            Normals              &  4.968  &  0.163  &  0.064 \\
            2D Edges             &  4.632  &  0.201  &  0.073 \\
            Scene Cls.           &  4.556  &  0.216  &  0.082 \\
            Obj. Cls.            &  4.699  &  0.236  &  0.091 \\
            2D Keypts.           &  4.440  &  0.337  &  0.100 \\
            Jigsaw               &  4.317  &  0.539  &  0.109 \\
            Blind                &  4.191  &  0.568  &  0.118 \\
            Autoenc.             &  4.113  &  0.577  &  0.127 \\
            Denoising            &  4.036  &  0.629  &  0.136 \\
            In-painting          &  4.061  &  0.693  &  0.145 \\
            2D Segm.             &  4.028  &  0.707  &  0.155 \\
            Pix-as-state         &  3.945  &  0.729  &  0.164 \\
            Rand. Proj.          &  3.782  &  0.754  &  0.173 \\
            Sem. Segm.           &  3.844  &  0.774  &  0.182 \\
            Vanish. Pts.         &  3.817  &  0.812  &  0.191 \\
            Occ. Edges           &  3.772  &  0.888  &  0.200 \\
            \hline
        \end{tabular}
    \end{center}
    \end{minipage}
  \vspace{0.5mm}
  \caption{\textbf{Features vs. scratch.} Gibson Navigation: Semantic features (both classification and segmentation) outperform scratch at timestep 420. Visual Exploration: timestep 480. Local Planning: timestep 620. P-values come from a Wilcoxon rank sum test adjusted for clustering effects. Features above the middle horizontal line are considered significant. The The false discovery rate (Q) is set to 20\%.}
  \label{fig:h2_cluster_analysis}
\end{figure*}

\subsection{Hypothesis III Experiments}

In this experiment we showed that geometric features are superior to semantic ones for exploration tasks, but that the opposite is true for navigation tasks. 

For each feature and for each task we train 10 agents from 10 random seeds. To evaluate the performance of each agent/seed combination we run it in the test environment for 100 random 100 episodes. We then use a cluster-effect-adjusted Wilcoxon rank-sum test to test whether there is a significant difference in the reward-per-episode between features.

\begin{figure}[H]\label{fig:h3_cluster}
  \centering
  \begin{minipage}{0.5\columnwidth}
  \footnotesize
  \begin{tabular}{ |c|c|c|c| } 
    \hline
    \textbf{Feature} & \textbf{Rew.} & \textbf{p-val.} &  $\mathbf{i/m}$ \\
    \hline
    \hline
    Sem. Segm.           &  6.553  &  0.0000  &  0.04 \\
    Scene Cls.           &  5.969  &  0.0001  &  0.08 \\
    Obj. Cls.            &  4.212  &  0.0004  &  0.12 \\
    Reshading            &  0.525  &  0.7824  &  0.16 \\
    3D Keypts.           &  -0.196  &  0.9460  &  0.20 \\
    \hline
    Distance             &  1.015  &  -  &  - \\
    \hline
  \end{tabular}
  \end{minipage}
\begin{minipage}{0.5\columnwidth}
  \footnotesize
  \begin{tabular}{ |c|c|c|c| } 
    \hline
    \textbf{Feature} & \textbf{Rew.} & \textbf{p-val.} &  $\mathbf{i/m}$ \\
    \hline
    \hline
    Distance             &  5.265  &  0.001  &  0.04 \\
    3D Keypts.           &  5.269  &  0.002  &  0.08 \\
    Reshading            &  5.072  &  0.011  &  0.12 \\
    Sem. Segm.           &  4.132  &  0.502  &  0.16 \\
    Scene Cls.           &  3.996  &  0.702  &  0.20 \\
    \hline
    Obj. Cls.            &  4.151  &  -  &  - \\
    \hline
  \end{tabular}
  \end{minipage}
  \vspace{0.5mm}
  \caption{\footnotesize{\textbf{Rank-reversal (Gibson).} Left: For navigation. Whether features are significantly better than distance estimation (the best feature for exploration) when used for navigation. The agents are evaluated after 420 updates (the end of training). Right: For exploration. Whether features are significantly better than object classification (the best feature for navigation) when used for exploration. The agents are evaluated after 480 updates (the end of training).}}
\end{figure}

\cleardoublepage

\section{Experimental setup details}

\subsection{Train/Test Splits}

\subsubsection{Gibson Train/Test Split}
In Gibson, we train in one building for both tasks (Beechwood, which is a medium-sized house) and evaluate the policies in 10 test environments given in the table below. The environments were selected for diversity. For exploration, large layouts on ground floors were selected (since the agent was trained on the ground floor). For navigation, generally open spaces were selected so the agent would need not have to take winding, complicated paths to the goal aside from very local (less than 3 m) obstacle avoidance (again, the reason for this choice being that the agent was trained in such a space).   

\begin{figure}
  \begin{minipage}{0.65\columnwidth}
  \centering
  \small
  \begin{tabular}{ |c|c|c| } 
    \hline
    \textbf{Task} & \textbf{Train Env.} & \textbf{Test Envs.} \\
    \hline
    \hline
    & & Ancor \\
    & & Aloha \\
    & & Corder \\
    & & Duarte \\
    Navigation & Beechwood & Eagan \\
    & & Globe \\
    & & Kemblesville \\
    & & Martinville \\
    & & Vails \\
    & & Area 1 (2D3DS) \\
    \hline
    & & Aloha \\
    & & Duarte \\
    & & Eagan \\
    & & Globe \\
    & & Hanson \\
    Exploration & Beechwood & Hatfield \\
    & & Kemblesville \\
    & & Martinville \\
    & & Sweatman \\
    & & Wiconisco \\
    \hline 
    
  \end{tabular}
  \end{minipage}
  \vspace{0.5mm}
  \caption{\footnotesize{\textbf{Train/Test Split (Gibson).} Environments chosen for train and test split in Gibson. Area 1 is from the Stanford 2D-3D-S dataset. \cite{2d3ds}} }
  \label{gibsontraintest}
\end{figure}

\subsubsection{Doom Texture Split}

Please see the files

\texttt{texture\_split/doom\_train\_textures.txt}

\noindent for a list of train textures and

\texttt{texture\_split/doom\_test\_textures.txt}

\noindent for a list of the test textures.

\subsection{Architectures}

\textbf{Pretrained features encoder:} For  all  tasks  we  modified  a  ResNet-50  encoder  with  no  average-pooling  and  replace  the  last  stride2 convolution with stride 1.  This gives us an output shape of $16 \times 16 \times 2048$;  we use a $3 \times 3$ convolution to transform the output to our final representation, which is of shape $16 \times 16 \times 8$.

We use frame-stacking for 4 frames and we share pretrained feature weights across each of these frames.

\textbf{Atari-net network:}

The atari-net network consists of three convolutional layers:

\begin{enumerate}
    \item Conv, 32 channel, 8x8 kernel, stride 4
    \item Conv, 64 channel, 4x4 kernel, stride 2
    \item Conv, 32 channel, 3x3 kernel, stride 1
    \item FC, output size = 512
\end{enumerate}

\textbf{Feature readout network:}

The feature readout network consists of one convolutional layer and two fully-connected layers:

\begin{enumerate}
    \item Conv, 32 channel, 4x4 kernel, stride 4
    \item FC, ouptut size = 1024
    \item FC, ouptut size = 512
\end{enumerate}

\subsection{Adapting Atari-Net for Mid-Level Features}

When \emph{features} are used as input in lieu of the pixel state, we apply a:

\begin{enumerate}
    \item Conv, 32 channel, 4x4 kernel, stride 2
\end{enumerate}

before the FC layers. This makes the intermediate representation immediately before flattening the same shape as the corresponding Atari-Net intermediate representation.

\subsection{Incorporating Side Information}

For \emph{Visual Exploration} when we incorporate the occupancy grid, we apply a separate Atari-Net tower to the occupancy grid, and concatenate its intermediate representation (before the FC layers) with the features/scratch intermediate representation before flattening and forwarding to the FC layers.\\

For \emph{Local Planning} when we incorporate the target vector, we expand each element of the target vector to an [H by W by 1] "image", and append each of these images to the back of the intermediate representation in the channel dimension before the last conv layer.

\subsection{Hyperparameters}

A complete list of hyperparameters can be found in the folder \texttt{hyperparameters}. For each task we conducted a grid search for the best hyperparameters for ``scratch''. We then used these hyperparameters for all features.

\subsection{Perception Module Distances}

We use the affinity matrix from~\cite{taskonomy2018}.

\subsection{Perception Module Architectures}

For each feature, we apply a:

\begin{enumerate}
    \item Conv, 32 channel, 4x4 kernel, stride 2
\end{enumerate}

We then concatenate and flatten the output and apply:

\begin{enumerate}
    \item FC, output size = 1024
    \item FC, output size = 512
\end{enumerate}

%-------------------------------------------------------------------------
{\small
\bibliographystyle{ieee}
\bibliography{egbib}
}
%-------------------------------------------------------------------------